\documentclass[10pt,twocolumn,letterpaper]{article}

 \usepackage[pagenumbers]{cvpr} % To force page numbers, e.g. for an arXiv version

\definecolor{cvprblue}{rgb}{0.21,0.49,0.74}
\usepackage[pagebackref,breaklinks,colorlinks,allcolors=cvprblue]{hyperref}

\def\confName{CVPR}
\def\confYear{2027}

\usepackage{graphicx, amsmath, amsfonts, array}
\newtheorem{Pro}{Proposition}

\usepackage{algorithm} 
\usepackage[noEnd=true,indLines=false]{algpseudocodex}

\title{ShapeNet: Shape Analysis of Functions and Curves via Orbit-valued Weights}

\author{Emmanuel Akeweje\\
Trinity College Dublin\\
Ireland\\
{\tt\small eakeweje@tcd.ie} \and Mimi Zhang\\
Trinity College Dublin\\
Ireland\\
{\tt\small mimi.zhang@tcd.ie}
}

\begin{document}
\maketitle

\begin{abstract}
The elastic Riemannian framework provides a principled geometry for the statistical analysis of functions and shapes, with the square-root velocity  function representation enabling efficient computation. However, incorporating elastic registration into neural networks presents two challenges. First, the optimal alignment must be updated in each forward pass, adding substantial computational cost. Second, backpropagation through the selected alignment can be difficult because small changes in the learned representations may cause abrupt switches between optimal alignment paths, making the alignment map nonsmooth. Building on recent functional neural networks, we introduce orbit-valued weights to extract phase-invariant features for shape analysis. To reduce computational overhead, we develop a parametric family of warping functions together with tailored initialization strategies. The isometric nature of the group action further allows us to decouple alignment from backpropagation. We demonstrate the advantages of ShapeNet over existing methods on open and closed curves and real neuroimaging data, showing improved accuracy in curve alignment and representation learning at a fraction of the computational cost. 
\end{abstract}    
\section{Introduction}
\label{sec:intro}

%Make sure that the Paper ID from the submission system is visible in the version submitted for review (replacing the ``*****'' you see in this document).
%If you are using the \LaTeX\ template, \textbf{make sure to update paper ID in the appropriate place in the tex file}.

%The space after \eg, meaning ``for example'', should not be a sentence-ending space. So \eg is correct, {\em e.g.} is not. The provided \verb'\eg' macro takes care of this.
%When citing a multi-author paper, you may save space by using ``et alia'', shortened to ``\etal'' (not ``{\em et.\ al.}'' as ``{\em et}'' is a complete word). If you use the \verb'\etal' macro provided, then you need not worry about double periods when used at the end of a sentence as in Alpher \etal.

Functional data typically exhibit two entangled sources of variability: \textit{phase}, differences in the timing of features such as peaks, and \textit{amplitude}, differences in their heights and shapes \citep[Section 4.7]{Srivastava2016FunctionalAS}. Phase is modeled by reparameterizing the domain, $f\mapsto f\circ h$, where $h$ is a warping function. Amplitude is the variation that remains after the functions are aligned by such warpings. Separating the two is a major challenge. In functional neuroimaging, temporal variation can obscure meaningful differences in neural responses and their relationships. For example, trial-to-trial latency jitter in EEG event-related potentials can blur averaged waveforms and attenuate their peaks \cite{MOLINA2024105531}. EEG responses can also stretch or compress with the duration of cognitive processes, challenging analyses that assume fixed-duration responses \cite{pnas.2214638119}. In resting-state fMRI, temporal misalignment between regional signals can affect correlation-based estimates of functional connectivity \cite{10.1093/cercor/bhab273}. Similar issues arise in computer vision. In object-contour matching and face recognition based on facial curves, differences in curve parameterization can obscure geometric similarity unless correspondences are established through reparameterization \citep{9151006}. In human-action recognition, variations in execution speed and local timing can misalign corresponding poses across otherwise similar action trajectories \citep{Lohit_2019_CVPR}.  These examples underscore the importance of accounting for phase variation when comparing functional observations, so that differences in timing or parameterization are not conflated with differences in amplitude or shape.

Elastic registration addresses this challenge through the square-root velocity (SRV) function representation, under which the alignment criterion can be expressed as an $L^2$ discrepancy \citep{srivastava2011registrationfunctionaldatausing}. A common computational procedure alternates between (1) estimating warping functions that align the SRV functions to a current template, often using dynamic programming, and (2) updating the template as the point-wise mean of the aligned SRV functions \citep{Marron2015FunctionalDA}. This procedure approximates a Fr{\'e}chet mean in the quotient space modulo reparameterization.

Recent neural approaches build on this framework by learning warping functions or approximating quantities invariant to phase variation. SrvfRegNet \cite{Chen_2021_CVPR} replaces numerical warp estimation with a neural network trained using an SRV function alignment loss. Its original implementation performs a single registration stage using the arithmetic mean of the input SRV functions as the reference. SrvfNet \cite{9522855} uses a variational autoencoder to model a distribution of warping functions and supports either a prescribed template or joint template estimation. DeepJAM \cite{Pham2025} extends neural registration to multivariate quasi-periodic functions. Using component-wise square-root slope functions, it estimates a shared warping function across each subject's channels and distinguishes local and global timing variation. In contrast, Hartman  \etal \cite{9522873}  learn to approximate elastic distances between function or curve pairs without explicitly predicting the aligning warps. Registration can also be integrated with a downstream learning task. DeepFRC \cite{ICLR2026_204810b4} jointly learns alignment and classification using an SRV-function based, class-aware alignment loss. The alignment functions are represented by Fourier coefficients and passed to a classifier, allowing the classification objective to influence warp estimation. SrvfRegNet, DeepJAM, and DeepFRC all use one-dimensional CNNs to extract features from the input functions (or their SRV transformations), and the extracted features are subject to constrained transformations to produce the warping functions.

Recently, several new neural network architectures have been developed to model functional data \citep{1007599, pmlr-v202-heinrichs23a}. A defining feature of these architectures is the use of functional layers whose weights are continuous functions or translation-invariant kernels, rather than scalar values. Universal approximation properties have been established for certain classes of functional neural networks \cite{STINCHCOMBE1999467, singh2025NeurIPS}. The learned functional weights (in the encoder) can further facilitate interpretation by identifying segments of the input functions most relevant to downstream tasks \cite{singh2026geometricfeaturelearningfunctional}. However, most existing functional neural networks do not explicitly account for phase variation in functional data. FAEclust \citep{singh2025NeurIPS} incorporates pairwise elastic distances between functions into its convex clustering objective to promote robustness to phase variation. However, this robustness pertains to the clustering results; the network does not enforce phase invariance in its learned representations.

Motivated by the universal approximation properties of functional neural networks, together with the treatment of reparameterized functions as equivalence classes in elastic shape analysis \citep{bauer2020intrinsicriemannianmetricsspaces}, we develop ShapeNet, a neural network architecture for functional data that is invariant to phase variation. Our main contributions are twofold: (1) a weight-orbit layer that guarantees invariance to domain reparameterization, and (2) a parametric warping approach that aligns functions at a speed comparable to fast DTW \citep{1367985.1367993}. The paper is organized as follows.  \cref{ShapeNet architecture} describes the weight-orbit layer and the overall network architecture. \cref{Training acceleration techniques} introduces the parametric warping function (PWF) approach and the initialization strategies tailored to it. \cref{Experiments} evaluate ShapeNet on simulated and real datasets across several tasks, including curve alignment, representation learning, and the analysis of neuroimaging data.

\section{ShapeNet architecture}\label{ShapeNet architecture}
\begin{figure}[!ht]
\centering
\includegraphics[width=\linewidth]{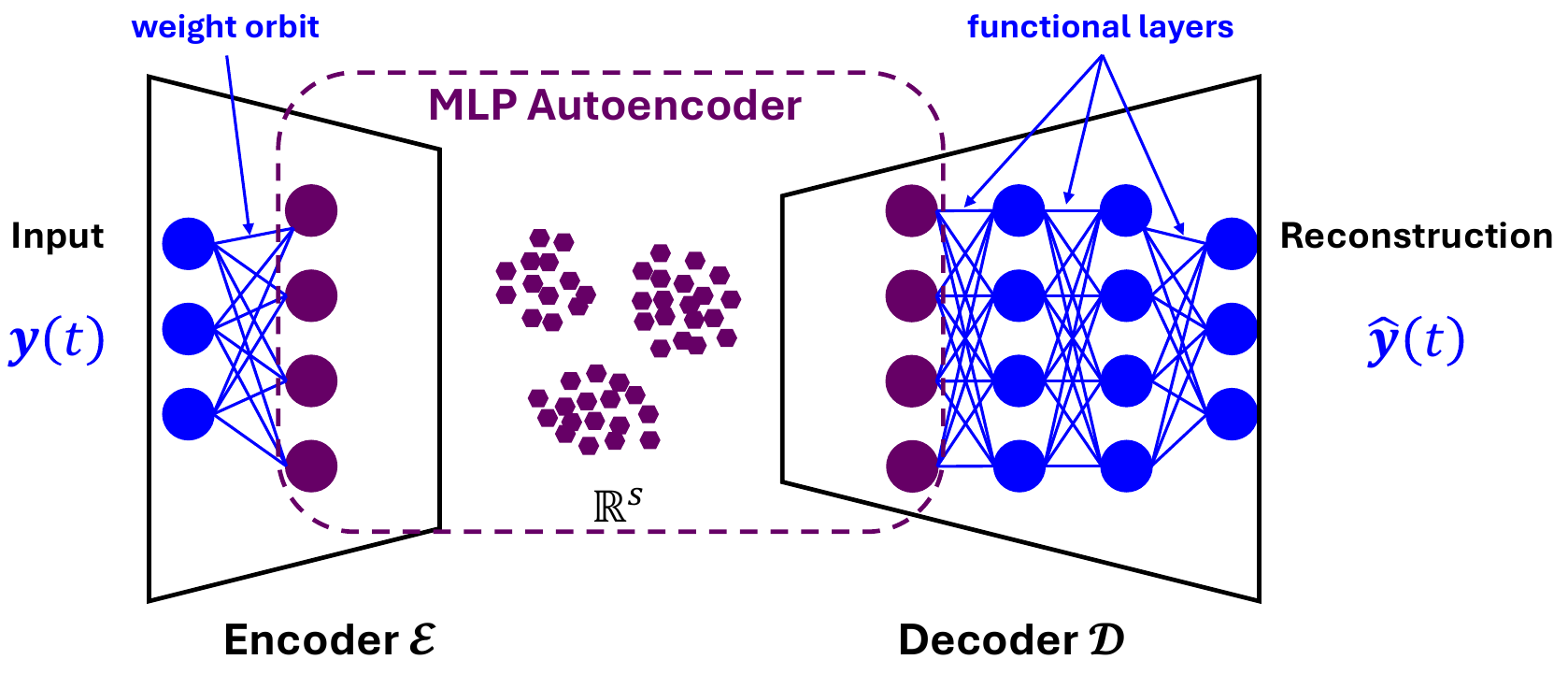}
\caption{Architecture of ShapeNet. The plum block is an MLP autoencoder; the blue components denote functional nodes/edges.}
\label{fig:ShapeNet_Arc}
\end{figure}

For ease of exposition, we assume that all functions are defined on the interval [0, 1]. ShapeNet comprises an encoder \(\mathcal{E}\) and a decoder \(\mathcal{D}\). The encoder maps a functional observation \(\pmb{y}\in\mathcal{H}([0,1],\mathbb{R}^p)\) to a latent vector representation \(\pmb{x}\in\mathbb{R}^s\), while the decoder maps \(\pmb{x}\) to a reconstructed function \(\hat{\pmb{y}}\in\mathcal{H}([0,1],\mathbb{R}^p)\) (\cref{fig:ShapeNet_Arc}). Here, $\mathcal{H}([0, 1], \mathbb{R}^p)$ is the separable Hilbert space of all square-integrable measurable functions that are defined on $[0, 1]$ and take values in $\mathbb{R}^p$.   The key architectural features underlying the effectiveness of ShapeNet are the \textit{orbit-valued weight layer} in the encoder and the \textit{phase-invariant reconstruction loss}.  Before explaining how ShapeNet separates shape information from phase variation, we briefly explain the SRV representation of a multi-dimensional function \cite{arxiv.1103.3817}, which provides the mathematical foundation for handling phase variation in functional data.

\subsection{The square-root velocity representation}
Let \(\{\pmb{y}_i=(y_i^1, \ldots, y_i^p)^\top\}_{i=1}^n\) denote \(n\) independent realizations of a \(p\)-dimensional random function \(\pmb{Y}=(Y^1, \ldots, Y^p)^\top\); that is, for each component \(1\leq d\leq p\), the scalar-valued functions  $\{y^d_1, \ldots, y^d_n\}$ are independent realizations of the random function $Y^d(t, \omega)$, $(t,\omega)\in[0, 1]\times\Omega$, defined on a probability space $(\Omega, \mathcal{F}, \Pr)$ and taking values in $\mathcal{H}([0, 1], \mathbb{R})$. In what follows, we suppress the dependence on \(\omega\) and write \(Y^d(t)\), or simply \(Y^d\), whenever no ambiguity arises. For each \(1\leq i\leq n\) and any $t\in[0, 1]$, \(\pmb{y}_i(t)=(y_i^1(t),\ldots,y_i^p(t))^\top\in\mathbb{R}^p\). We require that any sample function $\pmb{y}_i$ is absolutely continuous on $[0, 1]$; that is, the component functions $\{y_i^d\}_{d=1}^p$ are all absolutely continuous. 

Let $H=\{h: [0, 1]\rightarrow[0, 1]\}$ denote the set of all orientation-preserving diffeomorphisms: \(h(0)=0\), \(h(1)=1\) and \(h'(t)>0\) for all \(t\in[0, 1]\). Then for any $h\in H$, the composition $\pmb{y}_i\circ h$ is a re-parameterization of $\pmb{y}_i$. The SRV representation of $\pmb{y}_i$ is defined by
\[
\text{SRV}(\pmb{y}_i)(t) = \frac{\pmb{y}_i'(t)}{\sqrt{\|\pmb{y}_i'(t)\|_2}}, ~~ \text{if} ~~\|\pmb{y}_i'(t)\|_2\neq0,
\]
and \(\text{SRV}(\pmb{y}_i)(t)=0\), otherwise. Under the reparameterization \(h\), the SRV representation of \(\pmb{y}_i\circ h\) is
\[
\text{SRV}(\pmb{y}_i\circ h)(t)=\text{SRV}(\pmb{y}_i)(h(t))\sqrt{h'(t)}.
\]

\subsection{Weight orbit}

\begin{figure}[!ht]
\centering
\includegraphics[width=\linewidth]{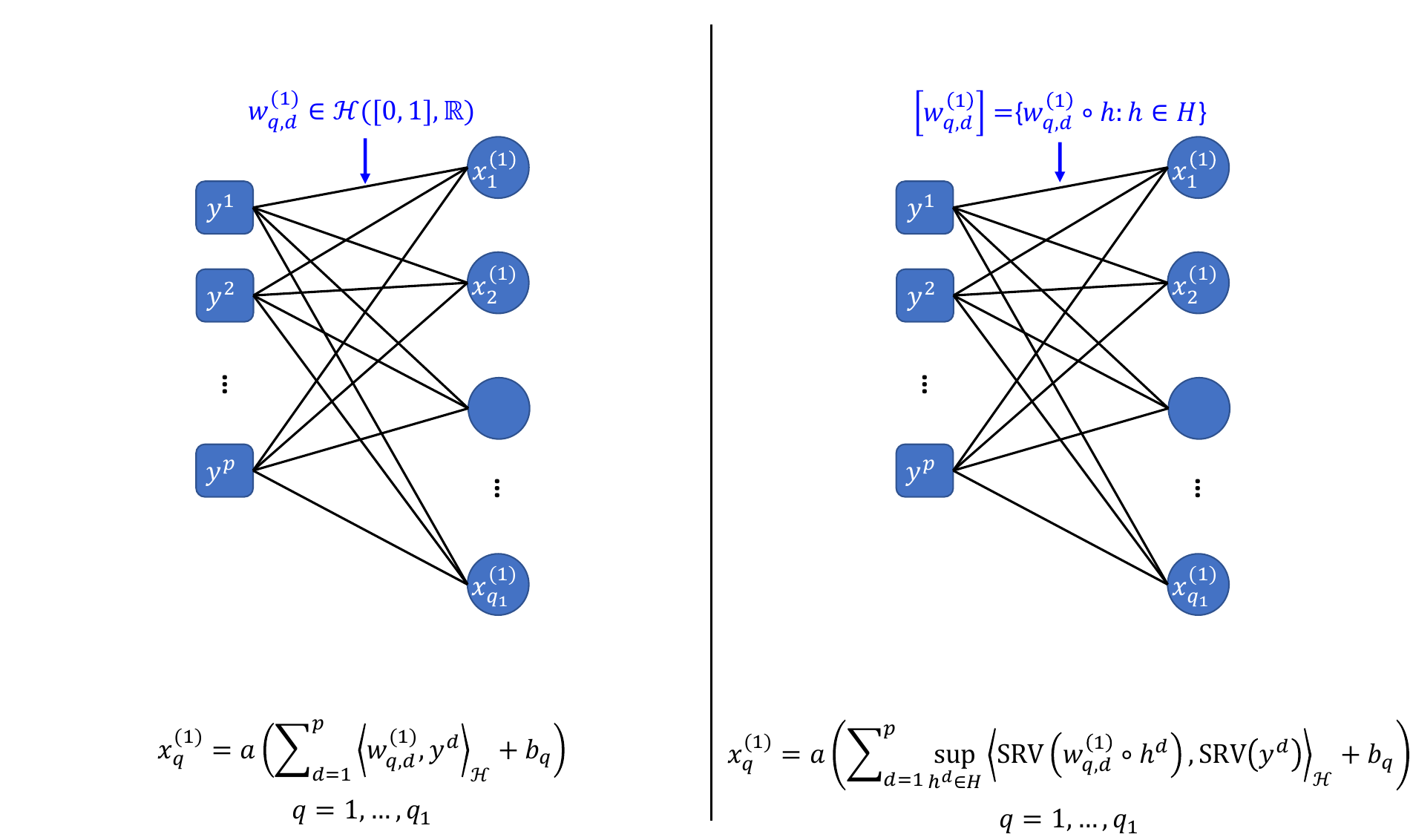}
\caption{The differences between the functional layers in FAEclust (left) and ShapeNet (right). The functional weights in the encoder of FAEclust are learned projection functions, and phase variation is handled through SRV-based similarity in the subsequent clustering step. ShapeNet directly removes phase variation by learning a phase-invariant projection.}
\label{fig:alignment-layer}
\end{figure}

In line with the architecture of FAEclust, ShapeNet contains four functional layers: one at the input of the encoder and three at the output of the decoder. Its phase-invariant capability relies on the encoder’s functional layer, which accounts for reparameterization when extracting features from the input functions. \cref{fig:alignment-layer} compares this layer (right) with its counterpart in FAEclust (left) and highlights the modification that enables ShapeNet to accommodate phase variation.

In FAEclust, the feedforward equation for the first functional layer is
\[
x^{(1)}_q=a( \sum_{d=1}^p  \langle w_{q, d}^{(1)}, y^d \rangle_{\mathcal{H}}+ b_q),~~~~ q=1, \ldots, q_1,
\]
where $a$ is a non-linear activation function, the $b_q$'s are bias parameters, and the weights \(w_{q,d}^{(1)}\in\mathcal{H}([0,1],\mathbb{R})\) are learnable functions rather than scalar-valued parameters. For each component \(d\), an orthogonality regularization is imposed on the collection of weight functions  $\{w_{q, d}^{(1)}\}_{q=1}^{q_1}$. This encourages different neurons to extract complementary, rather than redundant, features from the corresponding input function \(y^d\), thereby providing subsequent layers with a more diverse representation. From this perspective, each weight function acts as a projection direction, and the inner product $\langle w_{q, d}^{(1)}, y^d \rangle_{\mathcal{H}}$ measures how strongly the input function $y^d$ aligns with the feature represented by $w_{q, d}^{(1)}$.

To account for phase variation in the input functions, ShapeNet replaces this standard inner product with
\begin{equation}\label{eq:weight_orbit}
    \sup_{h\in H}\langle\text{SRV}( w_{q, d}^{(1)}\circ h), ~\text{SRV}(y^d)\rangle_{\mathcal{H}}.
\end{equation}
Note that, in Eq. \eqref{eq:weight_orbit}, the supremum is evaluated independently for each shape template $w_{q, d}^{(1)}$.

This operation measures the maximum agreement between the SRV representations of \(w_{q, d}^{(1)}\) and \(y^d\) over all admissible reparameterizations \(h\in H\). Importantly, this choice of similarity measure is directly linked to the elastic distance commonly used in functional shape analysis. In particular, for any two functions \(y, w \in\mathcal{H}([0, 1], \mathbb{R})\), we have
\begin{align*}
\left\|\text{SRV}(y)-\text{SRV}(w\circ h)\right\|_{\mathcal{H}}^2
&=\left\|\text{SRV}(y)\right\|_{\mathcal{H}}^2+\left\|\text{SRV}(w\circ h)\right\|_{\mathcal{H}}^2 \\
&\quad -2\left\langle \text{SRV}(y), \text{SRV}(w\circ h) \right\rangle_{\mathcal{H}}.
\end{align*}
For the SRV transformation, we have
\begin{align*}
\left\|\text{SRV}(w\circ h)\right\|_{\mathcal{H}}^2
&=
\int_0^1
\left|\text{SRV}(w)(h(t))\right|^2 h'(t)\,\mathrm{d}t \\
&=
\int_0^1
\left|\text{SRV}(w)(s)\right|^2\,\mathrm{d}s = \left\|\text{SRV}(w)\right\|_{\mathcal{H}}^2,
\end{align*}
where the substitution \(s=h(t)\) has been used. Therefore, the first two terms are independent of \(h\), and
\[
\hspace{-0.2cm}
\sup_{h\in H}\langle\text{SRV}(y),\text{SRV}(w\circ h)\rangle_{\mathcal{H}}
\Leftrightarrow
\inf_{h\in H}
\left\|\text{SRV}(y)-\text{SRV}(w\circ h)\right\|_{\mathcal{H}}^2.
\]
Hence, an optimal reparameterization \(h^\ast\) that maximizes the inner product also minimizes the elastic distance between \(y\) and \(w\). For functions with fixed SRV norms, a larger optimal inner product corresponds to a smaller elastic distance between \(y\) and \(w\). Geometrically, \(w\circ h^\ast\) is the point on the nonlinear reparameterization orbit \([w]\) that is closest to \(y\) under the Fisher-Rao metric. Although this is not an orthogonal projection onto a linear subspace, it is analogous to a nearest-point projection onto that orbit.

This relationship changes the interpretation of an edge in ShapeNet. In FAEclust, a weight function defines a single projection direction. In ShapeNet, however, a weight function represents an entire reparameterization orbit \([w_{q, d}^{(1)}]=\{w_{q, d}^{(1)}\circ h, h\in H\}\). The supremum $\sup_{h\in H} \langle \text{SRV}(w_{q, d}^{(1)}\circ h), \text{SRV}(y^d) \rangle_{\mathcal{H}}$  compares the input component $y^d$ with every representative in the orbit and selects the one that provides the strongest agreement in the SRV space. Consequently, \(w_{q, d}^{(1)}\) is more naturally interpreted as a \textit{shape template} defined up to reparameterization, and the corresponding edge represents a \textit{weight orbit} rather than a single weight function.

\paragraph{Decouple alignment from backpropagation. } In our code implementation, the input $y^d$ is registered to the shape template $w_{q, d}^{(1)}$: 
\begin{equation*}
x^{(1)}_q=a( \sum_{d=1}^p  \sup_{h\in H} \langle \text{SRV}(w_{q, d}^{(1)}), \text{SRV}(y^d\circ h) \rangle_{\mathcal{H}}+ b_q).    
\end{equation*}
Note again that the supremum is evaluated independently for each shape template $w_{q, d}^{(1)}$. Theoretically, registering the input to the shape template or the shape template to the input yields the same optimal similarity, provided that \(H\) is a group and its action on the SRV space is isometric. This equivalence follows because every admissible reparameterization \(h\) has an inverse \(h^{-1}\). Computationally, however, registering the input to the shape template is preferable during training. It leaves the learned shape template unwarped and provides a more direct gradient path from \(x^{(1)}_q\) to \(w_{q, d}^{(1)}\), avoiding the additional dependence introduced by composing the shape template with the optimized reparameterization. This leads to more stable and effective gradient propagation.

\subsection{Phase-invariant reconstruction}
When training ShapeNet, the Euclidean reconstruction loss is no longer suitable; instead, the phase-invariant elastic distance provides a more meaningful measure of reconstruction error. Formally, the reconstruction loss is defined as: \(L_r=\frac{1}{n}\sum_{d=1}^p \sum_{i=1}^n d_{\text{ela}}(y_i^d, \hat{y}_i^d)\), where
\begin{equation}
\label{supremum_reconstruction}
d_{\text{ela}}(y_i^d, \hat{y}_i^d)= \inf_{h\in H}\|\text{SRV}(y^d_i\circ h)-\text{SRV}(\hat{y}^d_i)\|_{\mathcal{H}}.
\end{equation}
This formulation allows the optimal warping function to differ across functional dimensions. 

When phase variation is assumed to be shared across all dimensions, a common warping function is applied to the entire multivariate functional observation. The reconstruction loss then becomes \(L_r= \frac{1}{n}\sum_{i=1}^n d_{\text{ela}}(\pmb{y}_i, \hat{\pmb{y}}_i)\), where
\begin{equation}
\label{reconstruction_shared_warping}
d_{\text{ela}}(\pmb{y}_i, \hat{\pmb{y}}_i)= \inf_{h\in H} \|\text{SRV}(\pmb{y}_i\circ h)-\text{SRV}(\hat{\pmb{y}}_i)\|_{\mathcal{H}}.
\end{equation}
Here, the same reparameterization \(h\) is used for all components of \(\pmb{y}_i\), thereby preserving their cross-dimensional phase correspondence.

\paragraph{Training objective.}
Following FAEclust, we include a roughness penalty on all functional weights and biases in the training objective: \(\mathcal{L} = L_r + \lambda_w L_w,\) where \(L_w\) penalizes functional weights and biases with high curvature, and \(\lambda_w \geq 0\) controls the strength of the penalty. Each functional weight and bias is expressed as a linear combination of predetermined, continuously differentiable basis functions (e.g., B-splines). We therefore implement the roughness penalty as an \(\ell_1\) penalty on the vector of basis-expansion coefficients.

\section{Training acceleration techniques}\label{Training acceleration techniques}

\subsection{Parametric warping function (PWF)}
Solving the optimization problems in \cref{eq:weight_orbit,supremum_reconstruction} by conventional dynamic programming is computationally demanding, as the optimal alignments must be recomputed repeatedly during the forward passes of network training. To improve computational efficiency, we represent the warping group \(H\) using a finite-dimensional parametric family. This formulation replaces the dynamic-programming search over discretized warping paths with an optimization over the unconstrained parameters \(\theta_1,\ldots,\theta_m\), thereby enabling more efficient network training.

Let \(B_{k,m}(t)=\binom{m}{k}t^k(1-t)^{m-k},~~0\leq k\leq m,\) denote the Bernstein basis polynomials of degree \(m\) on \([0, 1]\). Consider unconstrained parameters $\pmb{\theta}=(\theta_1,\ldots,\theta_m)^\top\in\mathbb{R}^m$ and define the strictly positive increments \(d_k=\exp(\theta_k)\),  \(k=1, \ldots, m\). The normalized increments are defined as 
\begin{equation*}
    \tilde{d}_k = d_k\times (\sum_{j=1}^{m}d_j)^{-1}.
\end{equation*} 
By construction, \(\tilde{d}_k>0\) and \(\sum_{k=1}^{m}\tilde{d}_k=1\). The Bernstein coefficients are defined through the cumulative sums
\[
c_0=0, ~~~~ c_k=\sum_{j=1}^{k}\tilde{d}_j, ~~~~ k=1,\ldots,m.
\]
The parametric form of the warping function is given by
\begin{equation}\label{parametric_warping_function}
    h_{\pmb{\theta}}(t) = \sum_{k=0}^{m}c_k B_{k,m}(t), ~~ t\in[0, 1].
\end{equation}
Since $c_0=0$ and $c_m=1,$ the endpoint conditions are satisfied exactly: $h_{\pmb{\theta}}(0)=0$ and $h_{\pmb{\theta}}(1)=1$. Furthermore, the consecutive coefficient differences satisfy $c_{k+1}-c_k= \tilde{d}_{k+1}>0.$ Using the standard differentiation identity for Bernstein polynomials, we obtain
\[
h_{\pmb{\theta}}'(t) = m\sum_{k=0}^{m-1} \left(c_{k+1}-c_k\right)B_{k,m-1}(t).
\]
Because the Bernstein basis polynomials are nonnegative on [0, 1] and all normalized increments are strictly positive,
\[
h_{\pmb{\theta}}'(t)=m\sum_{k=0}^{m-1}\tilde{d}_{k+1}B_{k,m-1}(t)>0, ~~ t\in[0,1].
\]
Therefore,  \(h_{\pmb{\theta}}\) is an orientation-preserving diffeomorphism from \([0, 1]\) onto itself, namely, \(h_{\pmb{\theta}}\in H\). Moreover, we have
\begin{Pro}\label{dense_parametric_family}
    For each \(m\in\mathbb{N}\), let \(H_m\) denote the family of functions \(\{\sum_{k=0}^{m}c_k B_{k,m}(t): t\in [0, 1]\}\), where \(c_0=0\), \(c_k=\sum_{j=1}^{k}\tilde{d}_j\), and 
    \[\tilde{d}_k = \frac{\exp(\theta_k)}{\sum_{j=1}^{m}\exp(\theta_j)},~~\pmb{\theta}\in\mathbb{R}^m.\] The union \(\bigcup_{m=1}^{\infty}H_m\) is dense, with respect to the \(C^1\) norm, in the set \(H\) of orientation-preserving \(C^1\)-diffeomorphisms from [0, 1] onto itself. That is, for every \(h\in H\) and every \(\delta>0\), there exist \(m\in\mathbb{N}\),  \(\pmb{\theta}\in\mathbb{R}^m\), and \(h_{\pmb{\theta}}\in H_m\) such that \(\|h-h_{\pmb{\theta}}\|_{C^1}<\delta\).
\end{Pro}
Proposition \ref{dense_parametric_family} justifies our parametric representation of the warping function: the family of all softmax-parameterized Bernstein polynomials of finite degree can approximate any orientation-preserving $C^1$-diffeomorphism of $[0, 1]$ arbitrarily closely in the $C^1$ norm. 

Optimization over the unconstrained parameters \(\theta_1,\ldots,\theta_m\) is straightforward, and Section \ref{Analytical gradients} of the Supplementary Material gives their analytical gradients. Note that the mapping \((\theta_1, \ldots, \theta_m) \mapsto (\tilde{d}_1, \ldots, \tilde{d}_m)\) has \(m\) input parameters, but the vector \((\tilde{d}_1, \ldots, \tilde{d}_m)\) lies on the simplex \(\sum_{j=1}^m\tilde{d}_j=1\), which has only \(m-1\) independent dimensions. Therefore, one parameter (direction) is redundant. We remove this redundancy by setting \(\theta_m=0\), and optimize only over \(\pmb{\theta}_{-m}=(\theta_1, \ldots, \theta_{m-1})\).

\subsection{Amortized initialization}
\paragraph{Multi-initialization via the Beta function } The optimization problems in \cref{eq:weight_orbit,supremum_reconstruction,reconstruction_shared_warping} are highly multimodal; therefore, a natural strategy is to use \textit{multiple initializations} of \(\pmb{\theta}_{-m}\). We note that the morphology of the Bernstein warping function \(h_{\pmb{\theta}}(t)\) in \cref{parametric_warping_function} is governed by the coefficient differences \(\tilde{d}_k=c_k-c_{k-1}\). For example, increasing values of \(\tilde{d}_k\) produce a convex warping function that changes slowly at the beginning and rapidly near the end. To generate initial warping functions with distinct morphologies, including convex, concave, S-shaped, and inverse S-shaped forms, we use the regularized incomplete Beta function \(I_t(\alpha,\beta)\). Specifically, for given \(\alpha, \beta>0\), the Bernstein coefficients are initialized as
\[
c_k=I_{\frac{k}{m}}(\alpha, \beta), \qquad k=0,\ldots,m.
\]
The choices \(\alpha=\beta=1\), \(\alpha>1,~\beta=1\), and \(\alpha=1,~\beta>1\) produce identity, convex, and concave warping functions, respectively. In addition, \(\alpha>1,~\beta>1\) produces an S-shaped function, whereas \(0<\alpha<1\) and \(0<\beta<1\) produce an inverse S-shaped function. We select multiple pairs of $(\alpha, \beta)$, covering the cases described above, and the corresponding multiple initial values of \(\pmb{\theta}_{-m}\) can then be calculated directly from the generated coefficients \(\{c_k\}_{k=1}^m\).

\begin{figure}[!t]
\centering
\includegraphics[width=\linewidth]{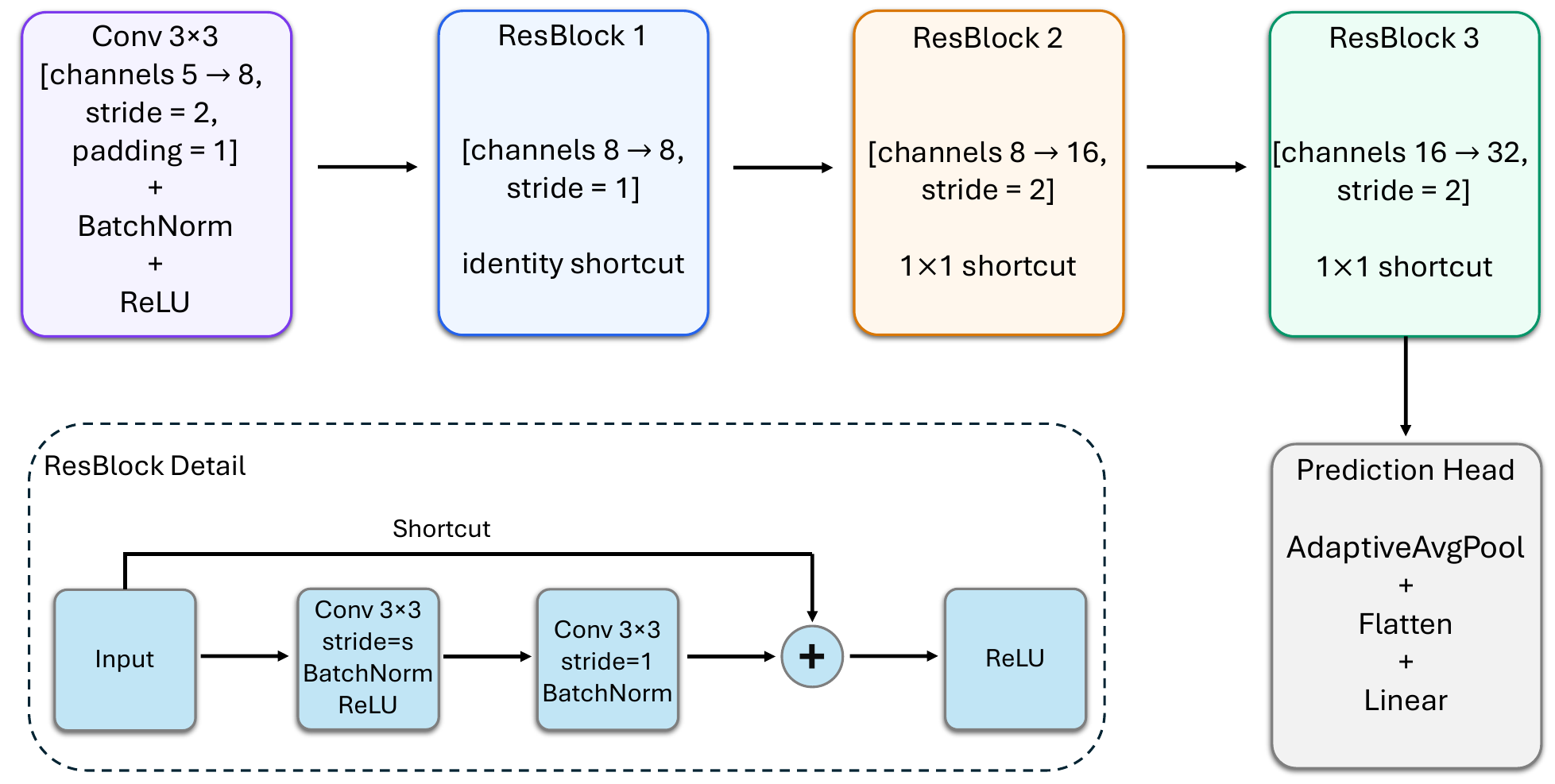}
\caption{Architecture of the initialization amortizer. The input is a five-channel pairwise time-matching tensor: \(\pmb{Y}\in\mathbb{R}^{5\times T_c\times T_c}\). The convolutional stem is bias-free. Within each residual block, the shortcut is the identity when the input and output dimensions agree; otherwise, a bias-free \(1\times1\) convolution with stride \(s\) is used (e.g., in ResBlock 3, stride=2). In the prediction head, adaptive average pooling maps the \(32\times \lceil T_c/8\rceil \times \lceil T_c/8\rceil\) feature map to \(32\times3\times3\). The resulting feature map is flattened into a 288-dimensional vector and passed through a linear layer to predict \(\pmb{\theta}_{-m}\). No output nonlinearity is applied.}
    \label{fig:cnn_amortizer}
\end{figure}

\paragraph{Single initialization via an amortizer}  For each function-reconstruction pair \((\pmb{y}_i,\hat{\pmb{y}}_i)\), we solve the optimization problem in \cref{reconstruction_shared_warping} using multiple initializations of \(\pmb{\theta}_{-m}\) and retain the solution, denoted by \(\pmb{\theta}_{-m}^{*i}\), that yields the minimal \(d_{\text{ela}}(\pmb{y}_i, \hat{\pmb{y}}_i)\). Given the input samples \(\{\pmb{y}_i\}_{i=1}^n\), the training procedure of ShapeNet will produce an auxiliary dataset of input-reconstruction pairs and the corresponding optimal coefficient vectors, \(\{(\pmb{y}_i, \hat{\pmb{y}}_i), \pmb{\theta}_{-m}^{*i}\}_{i=1}^n.\) When \(n\) is sufficiently large, we apply the above multi-initialization method only to a subset of the training samples, say \(\{\pmb{y}_i\}_{i=1}^{n_1}\), where $n_1<n$.  We then use the produced auxiliary dataset \(\{(\pmb{y}_i, \hat{\pmb{y}}_i), \pmb{\theta}_{-m}^{*i}\}_{i=1}^{n_1}\) to train an initialization amortizer $\mathcal{A}$ (\cref{fig:cnn_amortizer}), an auxiliary deep neural network that predicts the optimal parameter vector \(\pmb{\theta}_{-m}^{*}\) given the input \((\pmb{y}, \hat{\pmb{y}})\). For each remaining sample, it provides $\mathcal{A}(\pmb{y}_i, \hat{\pmb{y}}_i)= \hat{\pmb{\theta}}_{-m}^{*i}$, $i=n_1+1, \ldots, n$,  which serves as an informed initialization of \(\pmb{\theta}_{-m}\) and avoids the computational cost of the multi-initialization procedure. We detail the architecture of this initialization amortizer below.

For an input function $\pmb{y}\in \mathcal{H}([0,1],\mathbb{R}^p)$ and its reconstruction \(\hat{\pmb{y}}\), we evaluate their SRV functions \(\text{SRV}(\pmb{y})\) and \(\text{SRV}(\hat{\pmb{y}})\) on a common grid of \(T_c\) time points: \(\{t_1, \ldots, t_{T_c}\}\). Then from each function pair \((\pmb{y},\hat{\pmb{y}})\), we construct a five-channel time-time representation. In particular, at each location \((t_j, t_k)\), the five channels are \((s_{jk}, d_{jk}, t_j, t_k, t_j-t_k)\), where $s_{jk}$ measures the similarity between the two SRV vectors $\text{SRV}(\pmb{y})(t_j)\in \mathbb{R}^p$ and $\text{SRV}(\hat{\pmb{y}})(t_k)\in \mathbb{R}^p$, and \(d_{jk}\) measures the corresponding distance between them. Each function-reconstruction pair is therefore represented by a tensor \(\pmb{Y}\in \mathbb{R}^{5\times T_c\times T_c}\), which encodes the pairwise matching information between all sampled time points of the two functions. The desired warping approximately appears as a path through this matrix: \(t_k=h(t_j)\). The architecture of the amortizer is detailed in \cref{fig:cnn_amortizer}. Note that, with an appropriate similarity measure $s_{jk}$, an initialization amortizer can likewise be trained for the problems in \cref{eq:weight_orbit,supremum_reconstruction}.

\subsection{Time and memory complexity} 
Consider $n$ curves, each with $p$ channels observed at $T$ time points, with each warp represented by a Bernstein polynomial of degree $m$. Let  $q_1$ be the number of hidden nodes  in the weight-orbit layer, $s$ the number of initializations per shape template $w_{q, d}^{(1)}$, and  $l$ the maximum number of optimization steps. Each optimizer step for a single candidate warp costs $O(mpT)$, and there are $nq_1s$ candidates. The total registration cost of the forward pass in the weight-orbit layer is therefore $O(nq_1slmpT)$. The candidates are independent and can be processed in parallel mini-batches, while the $l$ updates for each candidate must be performed sequentially. For a mini-batch of $n_b$ curves, the peak working memory is $O(n_bq_1sp(T+m)+Tm)$, where $Tm$  accounts for the Bernstein basis evaluated on the time grid.

\section{Experiments}\label{Experiments}
All real datasets used are publicly available. Preprocessing steps and implementation details for all methods are provided in the Supplementary Material.

\subsection{Curve alignment}
We first evaluate the PWF method for curve alignment on six simulation scenarios and compare it with five existing curve alignment methods. The simulation scenarios are described in detail in the Supplementary Material. \cref{fig:A1_chirp_mean_sd_revised} shows the alignment results for the first scenario (A1), a chirp whose oscillation frequency increases with time.
\begin{figure*}
  \centering
  \begin{subfigure}{\linewidth}
    \includegraphics[width=\linewidth]{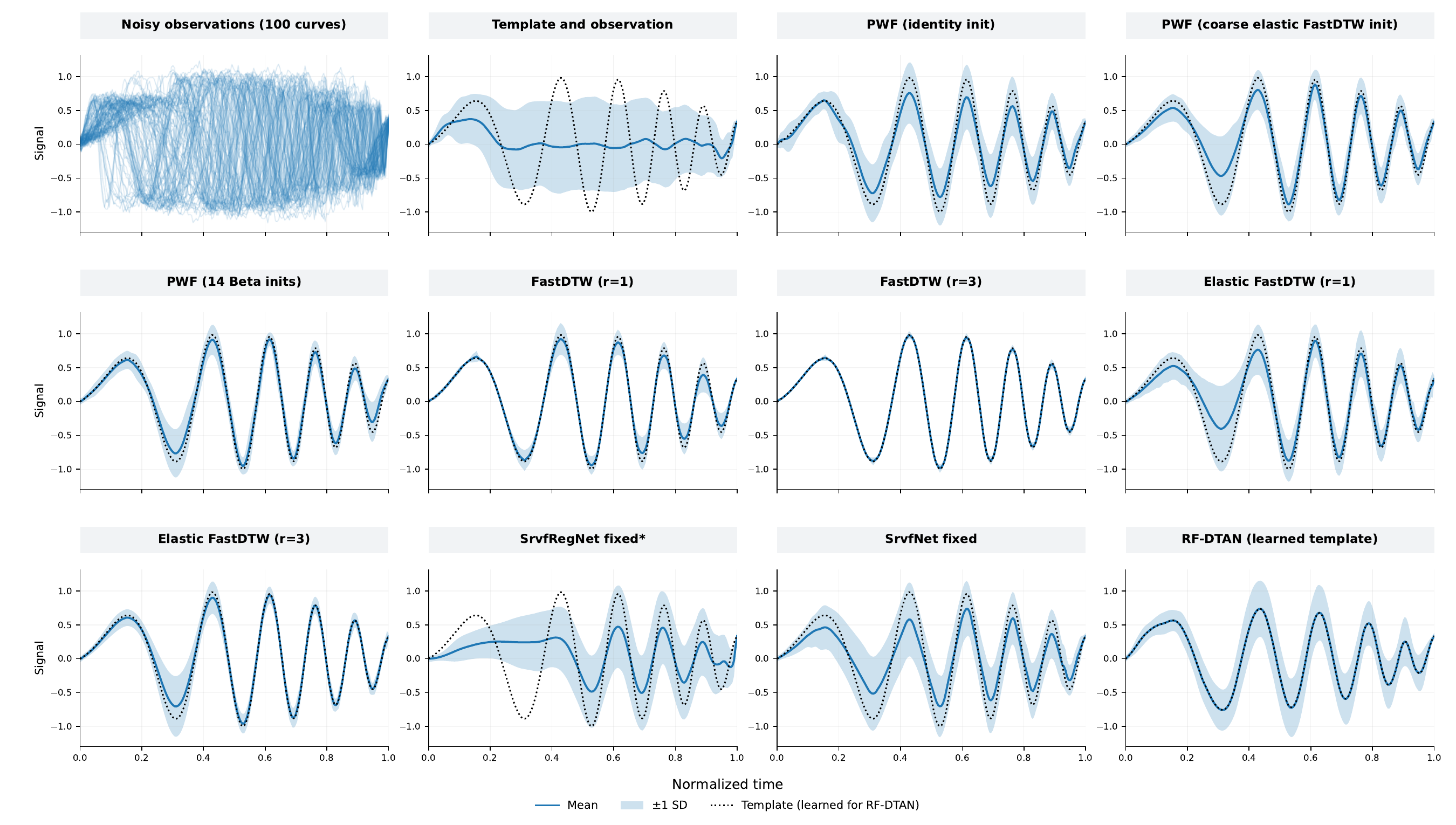}
  \end{subfigure}
  \caption{Alignment of 1000 simulated trajectories. The first panel in the top row shows a subset of 100 noisy trajectories. The remaining panels show the pointwise mean (solid line) and $\pm 1$ standard deviation (shaded band) of the curves, with the template overlaid as a dotted line. They cover in order: the unaligned curves; the proposed PWF initialized at the identity ($\alpha=\beta=1$), from a coarse FastDTW path ($r=1$), and from 14 Beta-function initializations; FastDTW applied to the smoothed curves and to their SRV functions (elastic FastDTW) with $r=1$ and $r=3$; and the SrvfRegNet, SrvfNet and RF-DTAN alignments.}
  \label{fig:A1_chirp_mean_sd_revised}
\end{figure*}

The panels are arranged in a $3\times 4$ grid and read left to right, top to bottom. The first panel shows 100 of the 1000 randomly generated noisy trajectories. Each remaining panel shows the pointwise mean (solid line) and $\pm 1$ standard deviation (shaded band) of the curves, with the template overlaid as a dotted line. The second panel shows the curves before alignment. Here phase variability flattens the mean and inflates the variance, so the structure of the template is lost. The remaining ten panels show the curves after alignment by different methods. We first consider three initialization strategies for our PWF. The third and fourth panels of the top row show PWF initialized at the identity warp, $\alpha=\beta=1$, and at the discrete alignment path produced by elastic FastDTW with search radius $r=1$. The first panel of the middle row uses 14 initial values of $(\alpha,\beta)$, corresponding to different warping morphologies. For comparison, the second and third panels of the middle row apply FastDTW directly to the curves with $r=1$ and $r=3$. The last panel of the middle row and the first panel of the bottom row apply FastDTW to the SRV functions of the curves (elastic FastDTW), again with $r=1$ and $r=3$. Finally, the last three panels of the bottom row show the alignments obtained by SrvfRegNet, SrvfNet and RF-DTAN \cite{Chen_2021_CVPR,9522855,weber2023regularization}. Since RF-DTAN learns its own template, the dotted line in its panel shows the learned template rather than the true one.

SrvfRegNet and SrvfNet perform worst. SrvfNet recovers the general structure but with damped amplitudes and wide bands throughout. SrvfRegNet  fails to align the first half of the curves and only partially recovers the later oscillation. RF-DTAN preserves the overall chirp pattern, but the later peaks and troughs are less pronounced in the aligned curve, and the wider SD band suggest greater variability among the aligned curves than for PWF. Among the PWF variants, the 14 initializations from the Beta function give the closest agreement with the template. The mean tracks all peaks and troughs, and the bands are relatively narrow, with only slight under-shooting at the first trough and the last peaks. The identity initialization recovers the overall shape but attenuates the amplitudes and leaves wider bands. Initialization from the coarse FastDTW path ($r=1$) improve the later oscillations but inherits that path's poor alignment of the first trough. This shows that PWF benefits from good initialization, and that exploring several warping morphologies is an effective way. FastDTW applied directly to the curves produces the tightest bands, especially with $r=3$. However, this tightness should not be read as better alignment. Because DTW on the raw curves minimized pointwise differences in function values, it is prone to pinching. Elastic FastDTW work with the SRV functions, which are designed to prevent pinching, so its bands are somewhat wider but its warps are more meaningful. Overall, elastic FastDTW behaves similarly to the PWF variants. However, a DTW path is not an orientation-preserving diffeomorphism. Even after interpolation, it belongs to at best the closure of $H$: the semigroup of absolutely continuous, nondecreasing maps $h$ with $h(0)=0$ and $h(1)=1$. By contrast, a PWF warp lies in $H$ by construction.

\begin{table}[!ht]
\centering
\caption{Computation time (in seconds) of each curve alignment method for scenario A1, for three combinations of the number of curves $n$ and the number of sampling points $T$. Entries are mean $\pm$ sample standard deviations over five runs. PWF was run on a single GPU, and the FastDTW variants on 24 CPU processes.}
\label{tab:a1-method-time}
\resizebox{\linewidth}{!}{%
\begin{tabular}{lccc}
\toprule
\textbf{Method}
& \shortstack{$n$=1000\\$T$=500}
& \shortstack{$n$=1000\\$T$=2000}
& \shortstack{$n$=10000\\$T$=500} \\
\midrule
FastDTW ($r$=1) & 0.94$\pm$0.02 & 9.43$\pm$0.07 & 2.30$\pm$0.02 \\
FastDTW ($r$=3) & 0.95$\pm$0.04 & 9.41$\pm$0.07 & 2.41$\pm$0.07 \\
Elas.FastDTW ($r$=1) & 1.51$\pm$0.04 & 10.68$\pm$0.20 & 6.83$\pm$0.17 \\
Elas.FastDTW ($r$=3) & 3.47$\pm$0.08 & 16.31$\pm$0.37 & 25.96$\pm$0.68 \\
PWF (identity init) & 2.31$\pm$0.04 & 2.23$\pm$0.05 & 2.57$\pm$0.13 \\
PWF (FastDTW init) & 3.09$\pm$0.03 & 5.72$\pm$0.31 & 10.60$\pm$0.52 \\
PWF (14 Beta inits) & 3.49$\pm$0.05 & 10.93$\pm$0.06 & 28.02$\pm$0.24 \\
\bottomrule
\end{tabular}%
}
\end{table}
\cref{tab:a1-method-time} reports the computation times for scenario A1, under three combinations of $n$ and $T$. The time taken by PWF with the identify initialization is almost insensitive to problem size. Although every run reaches the limit of 1,000 iterations, increasing $T$ fourfold changed its time from 2.31 to 2.23\,s, and increasing $n$ tenfold changed it from 2.31 to 2.57\,s. This is because PWF process all $n$ curves in a single batch on an RTX 5070 Ti GPU. In contrast, the time taken by the FastDTW variants, run on 24 CPU processes, grows sharply with $T$. These results reflect efficient GPU batching rather than a reduction in total computational work, and the relative timings depend on the hardware used. \cref{fig:A2_pwf_identity_convergence} shows that a reasonable alignment can be obtained well before 1,000 iterations. This motivates using only a few dozen PWF iterations when solving \cref{eq:weight_orbit,supremum_reconstruction,reconstruction_shared_warping}, which makes training the network efficient.

\subsection{Clustering and classification analysis} \label{cluster-analysis}

We first examine the clustering performance of ShapeNet by applying it to 12 real-world datasets and comparing its performance with that of existing functional neural networks designed for clustering: FAE \cite{hsieh2021functional}, VANO \cite{seidman2023variational}, and FAEclust \cite{singh2025NeurIPS}. The datasets comprise the optical motion-capture modality of the Berkeley Multimodal Human Action Database (BMHAD) \cite{ofli2013berkeley}, together with 11 datasets from the UEA \& UCR Time Series Classification Archives \cite{ueaucr}. They vary in sample size $n$, sequence length $T$, dimension $p$, and true number of clusters $K$; details are given in \cref{tab:arm-b-datasets}. Each dataset was split into stratified training, validation, and test sets. For all methods, hyperparameters were selected based on validation reconstruction loss, ensuring that model selection did not use true class labels; the full protocol is detailed in \cref{cluster_analysis_supplement} in the Supplementary Material. For ShapeNet, FAE, and VANO, we fitted K-means and Gaussian mixture models (GMMs) with diagonal covariance matrices to the training embeddings, using the known number of classes $K$ as the number of clusters. We then used the fitted clustering models to assign cluster labels to the held-out test embeddings. For FAEclust, we used its built-in convex-clustering procedure, which estimates the number of clusters from the training data, and assigned each test embedding to the nearest training-cluster centroid. Clustering quality was measured by the Adjusted Mutual Information (AMI) and Adjusted Rand Index (ARI), reported in \cref{tab:arm-b-ami,tab:arm-b-ari}. The tables use the abbreviations SL (SwedishLeaf), RS (RacketSports), and UWGL (UWaveGestureLibrary).

\cref{tab:arm-b-ami} reports the mean AMI over 20 random initializations, while \cref{tab:arm-b-ami-full,tab:arm-b-ari} additionally report standard deviations for AMI and ARI, respectively. ShapeNet achieves the highest mean AMI and ARI on 8 of the 12 datasets, including a tie with FAE (GMM) on ECG200. The choice of clustering method also matters. For ShapeNet, GMM improves both metrics over K-means on eight datasets. ShapeNet's reported standard deviations are generally small, especially on scalar-function datasets, although BMHAD shows greater variability, with an ARI standard deviation of 0.093 for GMM. These results indicate limited variability in the evaluation scores across initializations on most datasets. The FAEclust result for the Crop dataset is not reported because its computationally expensive convex-clustering step prevented training from completing within the 48-hour time limit.

\begin{table}[!ht]
\centering
\caption{Mean AMI, averaged over 20 random initializations.}
\label{tab:arm-b-ami}
\resizebox{\linewidth}{!}{%
\setlength{\tabcolsep}{2pt}
\begin{tabular}{lccccccc}
\toprule
& \multicolumn{2}{c}{FAE} & \multicolumn{2}{c}{VANO} & FAEcl. & \multicolumn{2}{c}{ShapeNet} \\
\cmidrule(lr){2-3} \cmidrule(lr){4-5} \cmidrule(lr){7-8}
\textbf{Data} & K-me. & GMM & K-me. & GMM &  & K-me. & GMM \\
\midrule
\multicolumn{8}{l}{\emph{Scalar functions}} \\
CBF & 0.514 & 0.386 & 0.388 & 0.382 & 0.310 & 0.563 & \textbf{0.570} \\
Crop & 0.533 & 0.497 & 0.509 & \textbf{0.564} & -- & 0.418 & 0.436 \\
ECG200 & 0.214 & \textbf{0.464} & 0.139 & 0.253 & 0.278 & \textbf{0.464} & 0.411 \\
Plane & 0.931 & 0.931 & 0.870 & 0.837 & 0.931 & \textbf{1.000} & \textbf{1.000} \\
SL & 0.428 & 0.551 & 0.444 & 0.534 & 0.192 & 0.556 & \textbf{0.602} \\
Symbols & 0.797 & 0.806 & 0.787 & 0.575 & 0.855 & \textbf{0.930} & \textbf{0.930} \\
Trace & 0.444 & 0.458 & 0.486 & 0.439 & \textbf{0.736} & 0.608 & 0.591 \\
\addlinespace
\multicolumn{8}{l}{\emph{Multi-dimensional functions}} \\
BMHAD & 0.780  & 0.719  & 0.864  & 0.786  &  0.312 & 0.848 & \textbf{0.908} \\
ERing & 0.596 & 0.614 & \textbf{0.952} & 0.931 & 0.647 & 0.596 & 0.635 \\
NATOPS & 0.483 & 0.531 & 0.702 & 0.756 & 0.464 & 0.714 & \textbf{0.807} \\
RS & 0.173 & 0.270 & 0.215 & 0.248 & 0.260 & 0.246 & \textbf{0.374} \\
UWGL & \textbf{0.736} & 0.457 & 0.598 & 0.628 & 0.243 & 0.256 & 0.308 \\
\bottomrule
\end{tabular}%
}
\end{table}

We next evaluate ShapeNet in a supervised setting by attaching a linear classification head to the ShapeNet encoder (hereafter referred to as ShapeNet-C). The resulting classifier is evaluated on the same 12 datasets and compared with three functional neural-network classifiers: FMLP \cite{rossi2005functional}, AdaFNN \cite{yao2021deep}, and FNN \cite{pmlr-v202-heinrichs23a}. Each model is trained on the training set, and both its hyperparameters and its retained checkpoint are selected using validation cross-entropy alone. The selected model is then evaluated once on the held-out test set using balanced accuracy, the mean of the per-class recalls, which accounts for class imbalance. Full details of the training and selection protocol are given in \cref{classification_analysis_supplement} in the Supplementary Material.

\cref{tab:arm-c-balanced-accuracy} shows that ShapeNet-C attains the highest balanced accuracy, either outright or tied, on seven of the 12 datasets. Among the scalar datasets, it is the sole best method on ECG200 and ties for best on CBF, Plane, and Symbols. Among the multi-dimensional datasets, ShapeNet-C performs best on BMHAD, ERing, and RS, while AdaFNN performs best on NATOPS and UWGL. Overall, these results show that the ShapeNet encoder is also effective for supervised classification of both scalar and multi-dimensional functional data, although no single method performs best on every dataset. \cref{tab:significance_test} in the Supplementary Material reports significance tests comparing ShapeNet and its supervised variant with the strongest competing clustering and classification methods, respectively.
\begin{table}[!ht]
\centering
\caption{Balanced classification accuracy.}
\label{tab:arm-c-balanced-accuracy}
\resizebox{0.85\linewidth}{!}{%
\setlength{\tabcolsep}{5pt}
\begin{tabular}{lcccc}
\toprule
\textbf{Data} & FMLP & AdaFNN & FNN & ShapeNet-C \\
\midrule
\multicolumn{5}{l}{\emph{Scalar functions}} \\
CBF     & 0.986 & \textbf{1.000} & \textbf{1.000} & \textbf{1.000} \\
Crop    & \textbf{0.705} & 0.598 & 0.687 & 0.604     \\
ECG200  & 0.675 & 0.500 & 0.725 & \textbf{0.775} \\
Plane   & 0.964 & 0.757 & \textbf{1.000}& \textbf{1.000} \\
SL      & 0.858 & 0.539 & \textbf{0.907} & 0.869 \\
Symbols & 0.942 & 0.832 & \textbf{0.987} & \textbf{0.987} \\
Trace   & 0.656 & 0.589 & \textbf{1.000} & 0.933\\
\addlinespace
\multicolumn{5}{l}{\emph{Multi-dimensional functions}} \\
BMHAD  & 0.982 & 0.936 & 0.955 & \textbf{0.991}  \\
ERing  & 0.890 & 0.935 & 0.929 & \textbf{0.958} \\
NATOPS & 0.907 & \textbf{0.926} & 0.648 & 0.852 \\
RS     & 0.832 & 0.856 & 0.832 & \textbf{0.878} \\
UWGL   & 0.833 & \textbf{0.955} & 0.894 & 0.819 \\
% GeitRec  & 0.605 & 0.478 & 0.541 & 0.553 \\ %(757322,10,101) with 5 classes
% ACDC  & 0.580 & 0.600 & 0.220 & 0.500 \\ %(150,8,129) with 5 classes
\bottomrule
\end{tabular}%
}
\end{table}

\subsection{Analysis of neuroimaging data}
We assess whether the weight-orbit layer can absorb temporal variability in real neural recordings, using the motor-imagery EEG recordings of BCI Competition IV Dataset 2a~\cite{tangermann2012review}. In these recordings, trial-to-trial differences in the latency and duration of task-related activity introduce phase variation that standard convolutional features do not explicitly model. We add a parallel weight-orbit branch to EEGNet \cite{lawhern2018eegnet} and concatenate the features from both branches before classification. We compare the resulting model with two baselines: EEGNet and FNN. To separate the effect of elastic alignment from that of the extra branch, we also include an ablation in which the branch is kept but alignment is disabled. Preprocessing and training details are given in \cref{Analysis of neuroimaging data_supp} of the Supplementary Material; classification and ablation results are reported in \cref{tab:eeg_results,tab:eeg_ablation}. 

\section{Conclusion}
\label{sec:conclusion}

We introduced ShapeNet, a functional neural network that incorporates elastic shape analysis through orbit-valued weights. Its weight-orbit layer extracts features invariant to domain reparameterization, while a phase-invariant reconstruction loss enables representation learning based on shape. To make repeated alignment computationally practical, we developed a Bernstein-polynomial warping family with approximation guarantees, supported by multiple initialization strategies and an initialization amortizer. The isometric action in SRV space also allows alignment to be decoupled from backpropagation through the learned templates. Experiments demonstrate effective curve alignment and competitive performance in both unsupervised and supervised learning. Nevertheless, a single smooth global warp can struggle to align sharp, closely spaced features, making PWF particularly sensitive to initialization in these settings. Future work will explore localized or piecewise warping families and extend the framework to other shape-preserving group actions, including rotations and uniform scaling of curves.

{
    \small
    \bibliographystyle{ieeenat_fullname}
    \bibliography{main}
}

% WARNING: do not forget to delete the supplementary pages from your submission % \input{sec/X_suppl}
 \clearpage
\setcounter{page}{1}
\setcounter{section}{0}
\maketitlesupplementary

\section{Proof of Proposition \ref{dense_parametric_family}}\label{P1_proof}
For any diffeomorphism \(h\in H\), its derivative \(h'\) is continuous and strictly positive on [0, 1]. Moreover, \(h(0)=0\) and \(h(1)=1\), and hence \(\int_0^1h'(t)dt=h(1)-h(0)=1\). For \(m\geq 1\), consider the Bernstein approximation of \(h'\):
\[
\tilde{h}_m(t)=\sum_{k=0}^{m}h'(\frac{k}{m}) B_{k,m}(t).
\]
By the Bernstein approximation theorem, \(\tilde{h}_m\) converges uniformly to \(h'\) as \(m \rightarrow \infty\):
\[
\|\tilde{h}_m-h'\|_\infty\rightarrow 0.
\]
Because \(h'\) is strictly positive, all coefficients \(h'(\frac{k}{m})\) are positive. It follows that \(\tilde{h}_m(t)>0\) on [0, 1]. Define \[a_m=\int_0^1 \tilde{h}_m(t)dt.\] Uniform convergence implies \[
a_m \rightarrow \int_0^1h'(t)dt=1.\] 
Now normalize \(\tilde{h}_m\) by setting \[\bar{h}_m(t)=\frac{\tilde{h}_m(t)}{a_m}.\] Then \(\bar{h}_m(t)>0\) and \[\int_0^1 \bar{h}_m(t)dt=1.\] 
Furthermore, \(\bar{h}_m\) converges uniformly to \(h'\):
\begin{align*}
    \|\bar{h}_m-h'\|_{\infty}&=\|\frac{\tilde{h}_m}{a_m}-h'\|_{\infty}\\
    &\leq \frac{1}{a_m} \|\tilde{h}_m-h'\|_{\infty}+|\frac{1}{a_m}-1|\|h'\|_{\infty}\rightarrow 0.
\end{align*}
Now define \(h_m(t)=\int_0^t \bar{h}_m(s)ds,\) then 
\[
h_m(0)=0, ~~h_m(1)=1,~~\text{and}~~h_m'(t)=\bar{h}_m(t)>0.
\]
Thus, \(h_m\) is an orientation-preserving polynomial diffeomorphism from [0, 1]  onto itself. We have for every \(t\in[0, 1]\), 
\begin{align*}
    |h_m(t)-h(t)|=|\int_0^t(\bar{h}_m(s)-h'(s))ds|\leq \|\bar{h}_m-h'\|_{\infty}.
\end{align*}
Consequently, \(\|h_m-h\|_{\infty}\rightarrow 0\). In addition, \[\|h_m'-h'\|_{\infty}=\|\bar{h}_m-h'\|_{\infty}\rightarrow 0.\] Therefore, \(\|h_m-h\|_{C^1}\rightarrow 0\).

It remains to show that each \(h_m\) belongs to the proposed parameterized family. Define 
\[
\tilde{d}_{k+1}=\frac{1}{(m+1)a_m}h'(\frac{k}{m})>0, ~~k=0, \ldots, m.
\]
Using \(\int_0^1 B_{k,m}(t)dt=\frac{1}{m+1}\), we obtain
\[
a_m=\int_0^1\tilde{h}_m(t)dt=\sum_{k=0}^{m}\int_0^1 h'(\frac{k}{m}) B_{k,m}(t)dt,
\]
and hence
\[
a_m=\frac{1}{m+1}\sum_{k=0}^{m}h'(\frac{k}{m}).
\]
By construction, we have
\[
\sum_{k=1}^{m+1} \tilde{d}_k=\sum_{k=0}^m\frac{1}{(m+1)a_m}h'(\frac{k}{m})=1.
\]
It follows that
\begin{align*}
    \bar{h}_m(t)&=\frac{1}{a_m}\sum_{k=0}^{m}h'(\frac{k}{m}) B_{k,m}(t)\\
    &=(m+1)\sum_{k=0}^{m}\tilde{d}_{k+1} B_{k,m}(t).
\end{align*}
Define \(c_0=0\), \(c_k=\sum_{j=1}^k \tilde{d}_j\) for \(k=1, \ldots, m+1\), and consider the Bernstein polynomial of degree \(m+1\):
\(
g_m(t)=\sum_{k=0}^{m+1}c_kB_{k,m+1}(t).
\)
Using the differentiation identity for Bernstein polynomials gives
\begin{align*}
    g_m'(t)&=(m+1)\sum_{k=0}^m (c_{k+1}-c_k)B_{k,m}(t)\\
    &=(m+1)\sum_{k=0}^m \tilde{d}_{k+1} B_{k,m}(t)=\bar{h}_m(t).
\end{align*}
From \(g_m(0)=c_0=0=h_m(0)\), we obtain that \(g_m=h_m.\)

Finally,  define \(\theta_k=\log(\tilde{d}_k)\) for \(k=1, \ldots, m+1\). Then 
\[
\frac{\exp(\theta_k)}{\sum_{j=1}^{m+1}\exp(\theta_j)}=\frac{\tilde{d}_k}{\sum_{j=1}^{m+1}\tilde{d}_k}=\tilde{d}_k.
\]
Thus, \(h_m=g_m\in H_{m+1}\). Since \(h_m\rightarrow h\) in the \(C^1\) norm, for every \(\delta>0\), there exists \(m\) sufficiently large such that \(\|h_m- h\|_{C^1}<\delta\). Therefore, \(\bigcup_{m=1}^{\infty}H_m\) is dense in \(H\) with respect to the \(C^1\) norm.

\section{Analytical gradients}\label{Analytical gradients}
Optimization over the unconstrained parameters \(\theta_1,\ldots,\theta_m\) is straightforward, as analytical gradients are available. For ease of exposition, write  \(q_w(t)\) for the SRV transformation of an absolutely continuous function $w$: \(q_w(t)=\text{SRV}(w)(t)\). For any shape template \(w\) and input function \(y\), define the similarity under the parametric
warping \(h_{\pmb{\theta}}\) by
\begin{align*}
    f(\pmb{\theta})&=\langle \text{SRV}(w), \text{SRV}(y\circ h_{\pmb{\theta}}) \rangle_{\mathcal{H}}\\
    &=\int_0^1 q_w(t) q_y(h_{\pmb{\theta}}(t))\sqrt{h_{\pmb{\theta}}'(t)}dt.
\end{align*}
Partial differentiation with respect to  \(\theta_i\) gives
\begin{align*}
    \frac{\partial f}{\partial\theta_i}&=\int_0^1 q_w(t) \times\\
    &\left[ q_y'(h_{\pmb{\theta}}(t))\sqrt{h_{\pmb{\theta}}'(t)}\frac{\partial h_{\pmb{\theta}}(t)}{\partial\theta_i}+\frac{q_y(h_{\pmb{\theta}}(t))}{2\sqrt{h_{\pmb{\theta}}'(t)}}\frac{\partial h_{\pmb{\theta}}'(t)}{\partial\theta_i}\right]dt.
\end{align*}
We below give the analytical expressions for \(\frac{\partial }{\partial\theta_i}h_{\pmb{\theta}}(t)\) and \(\frac{\partial }{\partial\theta_i}h_{\pmb{\theta}}'(t)\). Let
\[
d_j=\exp(\theta_j), ~~A=\sum_{k=1}^m d_k, ~~\tilde{d}_j=\frac{d_j}{A}.
\]
For \(1\leq i, j\leq m\), the quotient rule gives
\[
\frac{\partial \tilde{d}_j}{\partial \theta_i}=\frac{d_i\delta_{ij}A-d_jd_i}{A^2}=\tilde{d}_i(\delta_{ij}-\tilde{d}_j),
\]
where \(\delta_{ij}\) denotes the Kronecker delta. Then we have
\[
\frac{\partial c_k}{\partial \theta_i}=\sum_{j=1}^{k}\frac{\partial \tilde{d}_j}{\partial \theta_i}=\tilde{d}_i(\sum_{j=1}^{k}\delta_{ij}-\sum_{j=1}^{k}\tilde{d}_j)=\tilde{d}_i(1_{\{i\leq k\}}-c_k).
\]
Consequently,
\begin{align*}
 \frac{\partial h_{\pmb{\theta}}(t)}{\partial\theta_i}=\sum_{k=0}^{m}\frac{\partial c_k}{\partial\theta_i} B_{k,m}(t)&=\tilde{d}_i\sum_{k=0}^{m} (1_{\{i\leq k\}}-c_k) B_{k,m}(t)\\
 &=\tilde{d}_i[\sum_{k=i}^{m} B_{k,m}(t)-h_{\pmb{\theta}}(t)].
\end{align*}
Using the Bernstein differentiation identity,
\[
h_{\pmb{\theta}}'(t)=m\sum_{k=0}^{m-1} \tilde{d}_{k+1} B_{k,m-1}(t)=m\sum_{j=1}^{m} \tilde{d}_j B_{j-1,m-1}(t).
\]
Therefore,
\begin{align*}
\frac{\partial h_{\pmb{\theta}}'(t)}{\partial \theta_i}&=m\sum_{j=1}^{m} \frac{\partial \tilde{d}_j}{\partial \theta_i} B_{j-1,m-1}(t)\\
&=m\tilde{d}_i\sum_{j=1}^{m} (\delta_{ij}-\tilde{d}_j) B_{j-1,m-1}(t)\\
&=\tilde{d}_i\left[mB_{i-1,m-1}(t)-h_{\pmb{\theta}}'(t)\right].
\end{align*}

Note that the mapping \((\theta_1, \ldots, \theta_m) \mapsto (\tilde{d}_1, \ldots, \tilde{d}_m)\) has \(m\) input parameters, but the vector \((\tilde{d}_1, \ldots, \tilde{d}_m)\) lies on the simplex \(\sum_{j=1}^m\tilde{d}_j=1\), which has only \(m-1\) independent dimensions. Therefore, one parameter (direction) is redundant. We remove this redundancy by setting \(\theta_m=0\), and optimize only \((\theta_1, \ldots, \theta_{m-1})\), which gives 
\[
\tilde{d}_j=\frac{\exp(\theta_j)}{1+\sum_{k=1}^{m-1} \exp(\theta_k)}, ~~j=1, \ldots, m-1,
\]
and
\[
\tilde{d}_m=\frac{1}{1+\sum_{k=1}^{m-1} \exp(\theta_k)}.
\]
The above partial derivative with respect to \(\theta_i\), for \(i=1, \ldots, m-1\), remains the same.

\section{Supplementary experimental details}
ShapeNet is available at
\href{https://anonymous.4open.science/r/ShapeNet-03FE/}{\nolinkurl{anonymous-repo}}.

\subsection{Comparative study of alignment performance}
We consider six function registration scenarios, A1-A6.  In each scenario, we generate 1,000 curves observed at 200 equally spaced points on [0, 1]. Throughout, $\phi(t; \mu, \sigma)$ denotes a Gaussian density function with mean $\mu$ and variance $\sigma^2$, and $s(u)$ denotes the logistic sigmoid function. For A2 and A3, each curve was generated from a scenario-specific base function $b_i$ with randomly perturbed parameters. The registration template is a fixed, unwarped canonical version of the base function, whose parameter values are given below. The other scenarios have no random parameters, and hence the template is the base function itself.

\paragraph{A1: Chirp.}
The base function is an amplitude-modulated chirp whose oscillation frequency increases over time,
\[
b_i(t)=
\left[0.35+0.65\sin(\pi t)\right]
\sin\!\left[2\pi(1.2t+4t^2)\right].
\]

\paragraph{A2: Two peaks.}
The base function is a mixture of two Gaussian density functions:
\[
b_i(t)=H_{i1}\phi (t;0.32,W_{i1}) + 0.75H_{i2}\phi(t;0.69,W_{i2}),
\]
where $H_{i1},H_{i2}\sim U(0.85, 1.15)$, $W_{i1}\sim U(0.045, 0.065)$, and $W_{i2}\sim U(0.065, 0.09)$. The parameters for the template are $(H_{i1}, H_{i2})=(1, 1)$ and widths $(W_{i1}, W_{i2})=(0.055, 0.0775)$.

\paragraph{A3: Repeated peaks.} The base function repeats a five-bump motif three times over [0, 1]. Let $r(t)=3t-\lfloor 3t\rfloor$, which increases from 0 to 1 on each of the intervals [0, 1/3), [1/3, 2/3) and [2/3, 1). Then
\[
\begin{aligned}
b_i(t)=& C_i\big[
0.30\phi(r(t);0.18,0.055)
-0.24\phi(r(t);0.36,0.035)\\
&+1.25\phi(r(t);0.43,0.050)
-0.35\phi(r(t);0.51,0.035)\\
&+0.42\phi(r(t);0.70,0.095)\big],
\end{aligned}
\]
with $C_i\sim U(0.93,1.07)$. The parameter value for the template is $C_i=1$.

\paragraph{A4: Smooth transition.}
The base function is a single sigmoid transition,
\[
b_i(t)=s\!\left(\frac{t-0.50}{0.055}\right).
\]

\paragraph{A5: Plateau with a notch.}
The base function is a plateau formed by a rising and falling sigmoid, with a small Gaussian notch at its center,
\[
b_i(t)=
s\!\left(\frac{t-0.20}{0.025}\right)
-s\!\left(\frac{t-0.75}{0.045}\right)
-0.17\phi(t;0.50,0.025).
\]

\paragraph{A6: Lorenz trajectory.} The base function was taken from a trajectory of the Lorenz system
\[
\frac{dx}{dt}=\alpha(y-x), \frac{dy}{dt}=x(\rho-z)-y, \frac{dz}{dt}=xy-\beta z,
\]
with the standard parameter values $\alpha = 10, \rho = 28, \beta = 8/3$. The system was integrated numerically from the initial state (1, 1, 1). Because these parameter values produce chaotic dynamics, the resulting base function is an irregular, non-periodic, three-dimensional signal, which also severs as the template. Each observation is therefore a three-dimensional trajectory whose coordinates share a common warping function.

Phase variation was introduced through a random warping function $h_i$ generated as follows. We first generate a random smooth function $L_i$.  For A2 and A5, $L_i$ was a sum of two Gaussian density functions with random means; for the remaining scenarios, it was a random sum of three Fourier terms. We standardize $L_i$ to have mean 0 and standard deviation 1 over the sampling grid, giving $Z_i$, and then set
\[
q_i(t)=\min\!\left(1,\max\!\left(-1,s_iZ_i(t)\right)\right),
~
h_i(t)=
\frac{\int_0^t e^{q_i(u)}\,du}
     {\int_0^1 e^{q_i(u)}\,du}.
\]
The strength $s_i$ determines how far $h_i$ departs from the identity, and we used  $s_i=0.3$ (weak warping) for 500 curves and $s_i=1$ (strong warping) for the other 500 curves. The noisy observation was
\[
x_i(t)=f_i\!\left(h_i^{-1}(t)\right)+\epsilon_i(t),
\quad
\epsilon_i(t_j)\overset{\mathrm{iid}}{\sim}
N(0,0.035^2).
\]
Thus, an estimated alignment warp maps template time to observation time and acts through $x_i(\hat{h}_i(t))$.

Before alignment, all noisy trajectories were smoothed using B-splines without any roughness penalty.  The numbers of basis functions were 60 for A2, A4, and A5; 25 for A6; 80 for A1; and 120 for A3. Every alignment method received the same smoothed trajectories. The template was neither warped nor smoothed.

We below provide details on how each alignment method is configured.
\begin{itemize}
    \item For the PWF approach, we represent each warp by a Bernstein polynomial with degree $m=30$  and minimize its elastic distance to the fixed template. The optimization uses Adam with learning rate $0.05$ and parameters $(\beta_1, \beta_2)=(0.9,0.99)$, and runs for at most 1,000 iterations. Optimization stops early if the relative improvement in the objective remains below $10^{-5}$ for 20 consecutive iterations.  The 14 Beta initializations consist of the nine pairs $(\alpha, \beta)\in\{1,1.5,2\}^2$, together with $(3,2)$, $(2,4)$, $(4,4)$, $(0.75,0.75)$, and $(0.65,0.65)$. The elastic FastDTW initialization uses a search radius of 1 on a coarse grid of 32 time points and 10 CPU threads; subsequent PWF optimization used the full 200-point grid on CUDA. All initialization variants processed all 1,000 curves as a full batch.
    \item We included Euclidean-distance FastDTW \cite{salvador2007fastdtw,tanida2019fastdtw} with radii 1 and 3, and a Numba-optimized elastic-distance FastDTW implementation \cite{singh2025NeurIPS} with radii 1 and 3. These dynamic-programming methods aligned curves using 10 CPU processes. 
    \item We implemented two SRV-based neural network comparators, SrvfNet \cite{9522855} and SrvfRegNet \cite{Chen_2021_CVPR} both in their fixed-template form. SrvfNet provides such a variant, but the original SrvfRegNet does not support alignment to an externally specified template. We therefore adapt its elastic loss to use the prescribed template as the reference. Both networks are trained on CUDA with batch size 128 for at most 1,000 epochs, using 800 curves for parameter updates and the remaining 200 for validation. Training stops early if the validation loss does not improve for 50 epochs, with a minimum of 100 epochs. To compare all methods on the same data, the trained networks are then used to align all 1,000 curves, including those used for training and validation.
    \item RF-DTAN \cite{weber2023regularization} was trained with its inverse-consistency averaging error loss. Unlike the methods above, RF-DTAN neither uses SRV functions nor aligns to a prescribed template; instead, it learns the template jointly with the warps. Training follows the same protocol as for the SRV-based networks (data split, epoch limit, and early-stopping rule), except that the batch size is 64. As before, the trained network is then applied to all 1,000 curves for the final comparison.
\end{itemize}

\begin{figure*}
  \centering
  \begin{subfigure}{\linewidth}
    \includegraphics[width=\linewidth]{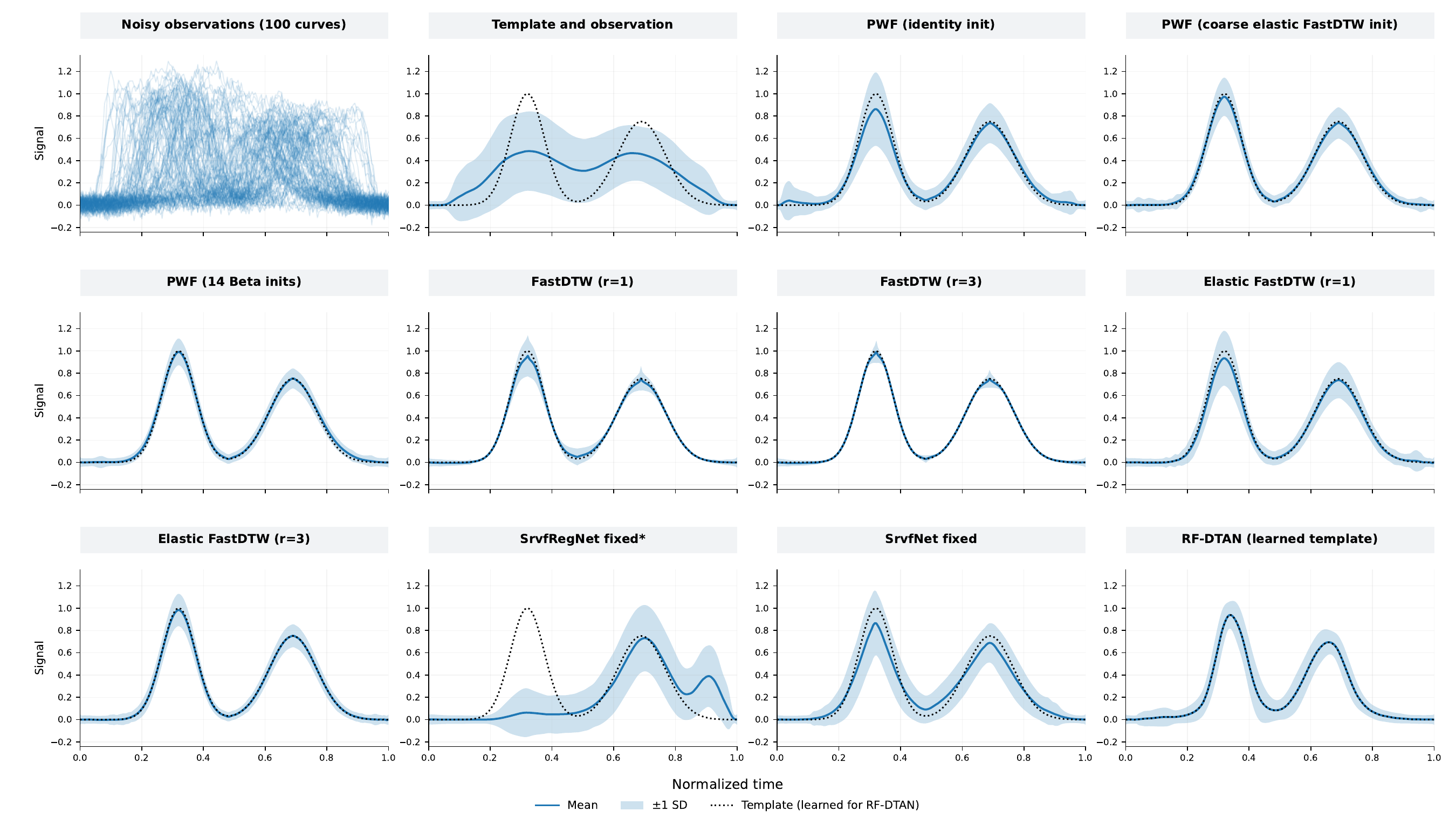}
  \end{subfigure}
\caption{Alignment of 1000 noisy trajectories simulated from A2.}
\label{fig:A2_peaks_mean_sd_revised}
\end{figure*}
\begin{figure*}
  \centering
  \begin{subfigure}{\linewidth}
    \includegraphics[width=\linewidth]{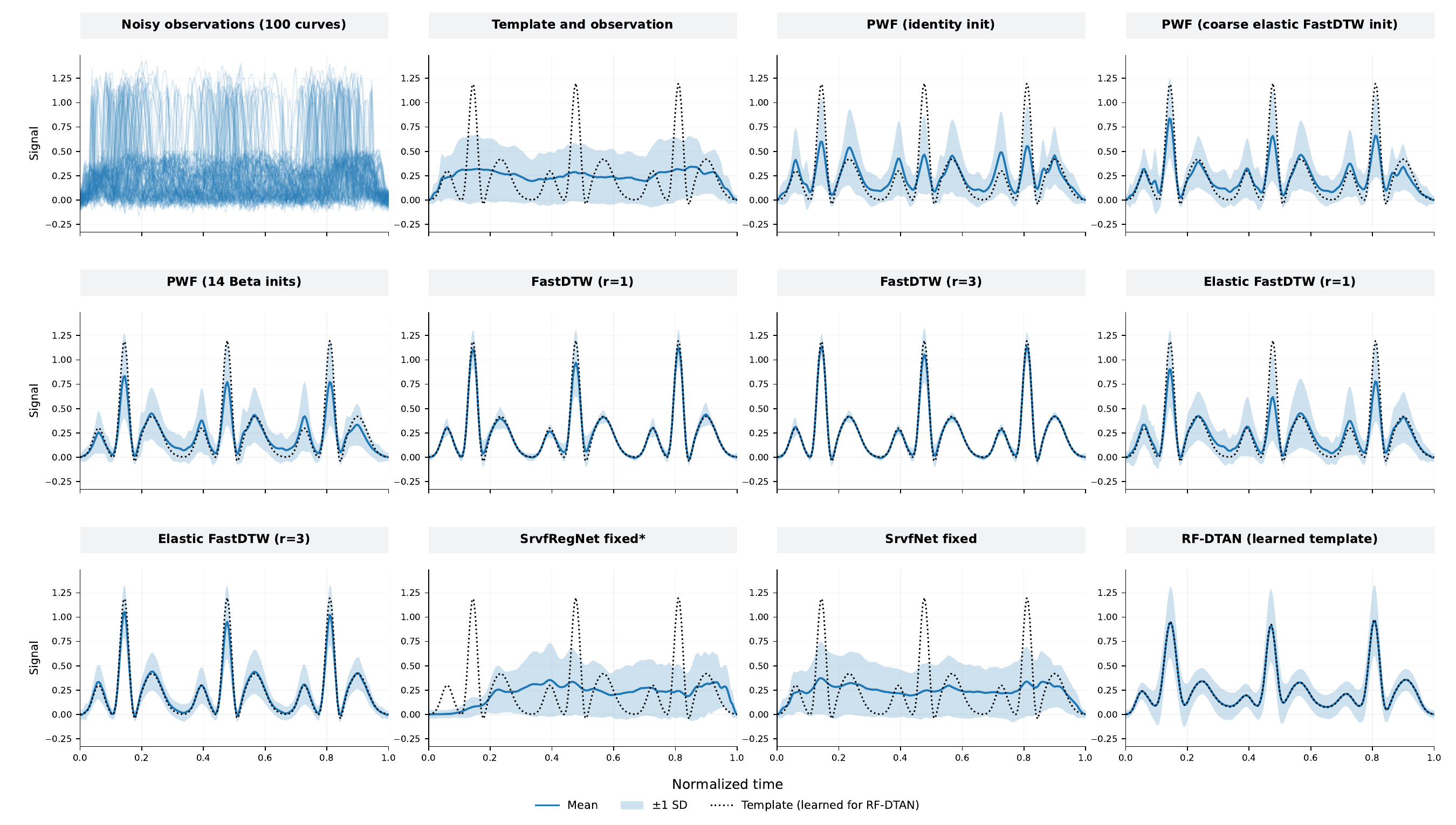}
  \end{subfigure}
\caption{Alignment of 1000 noisy trajectories simulated from A3.}
\label{fig:A3_periodic_mean_sd_revised}
\end{figure*}
\begin{figure*}
  \centering
  \begin{subfigure}{\linewidth}
    \includegraphics[width=\linewidth]{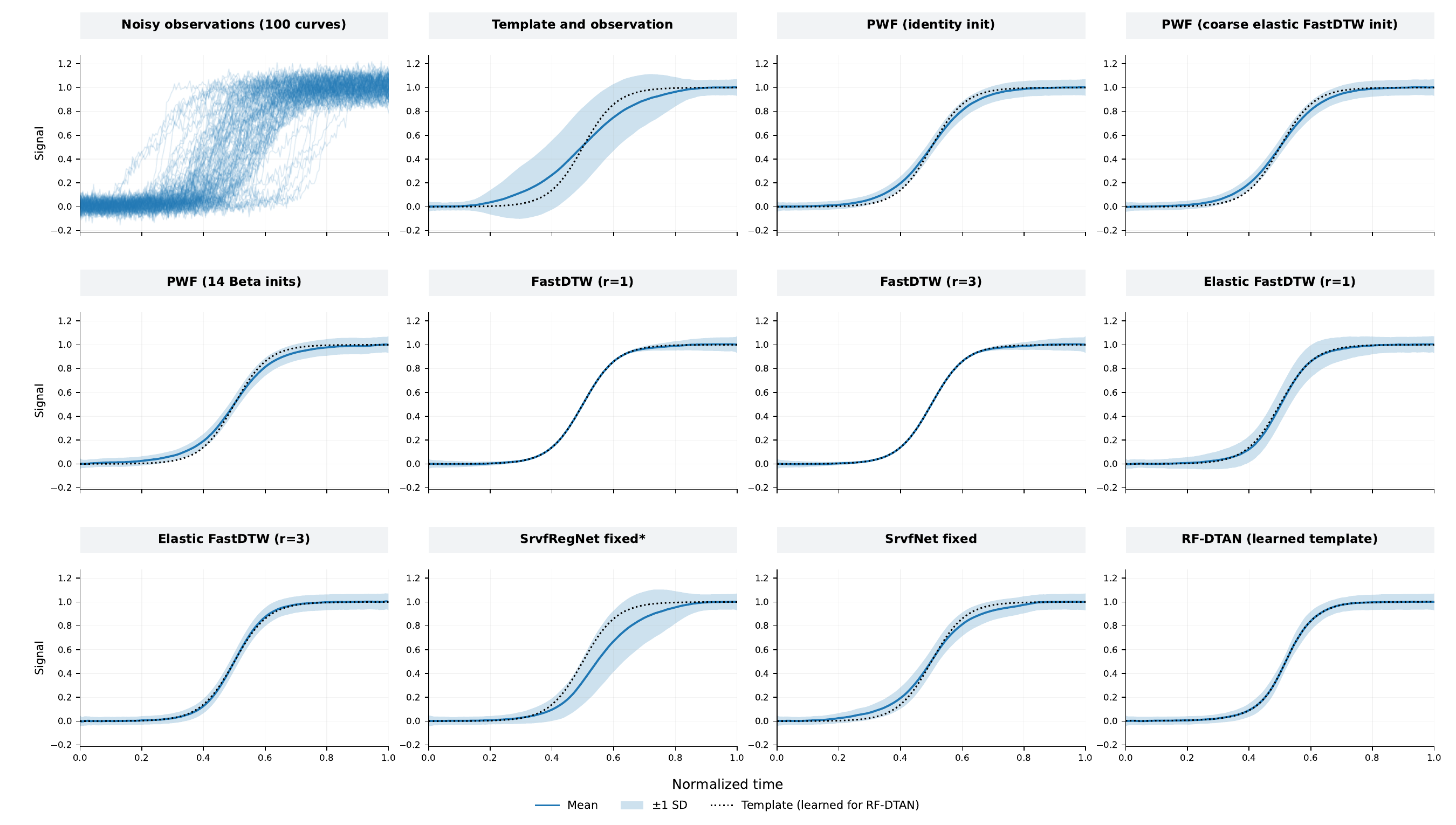}
  \end{subfigure}
\caption{Alignment of 1000 noisy trajectories simulated from A4.}
\label{fig:A4_transition_mean_sd_revised}
\end{figure*}
\begin{figure*}
  \centering
  \begin{subfigure}{\linewidth}
    \includegraphics[width=\linewidth]{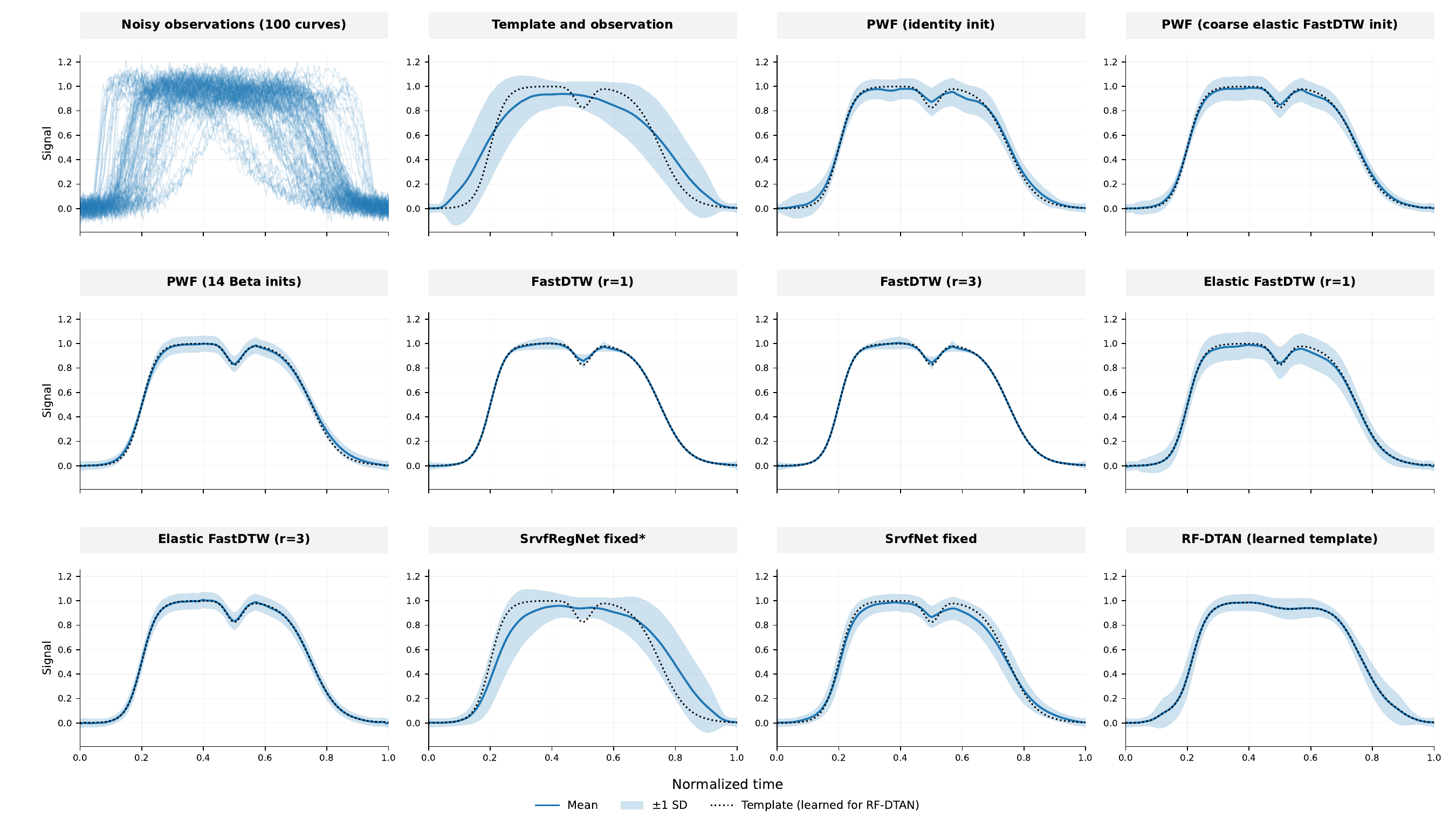}
  \end{subfigure}
\caption{Alignment of 1000 noisy trajectories simulated from A5.}
\label{fig:A5_plateau_mean_sd_revised}
\end{figure*}
\begin{figure*}
  \centering
  \begin{subfigure}{\linewidth}
    \includegraphics[width=\linewidth]{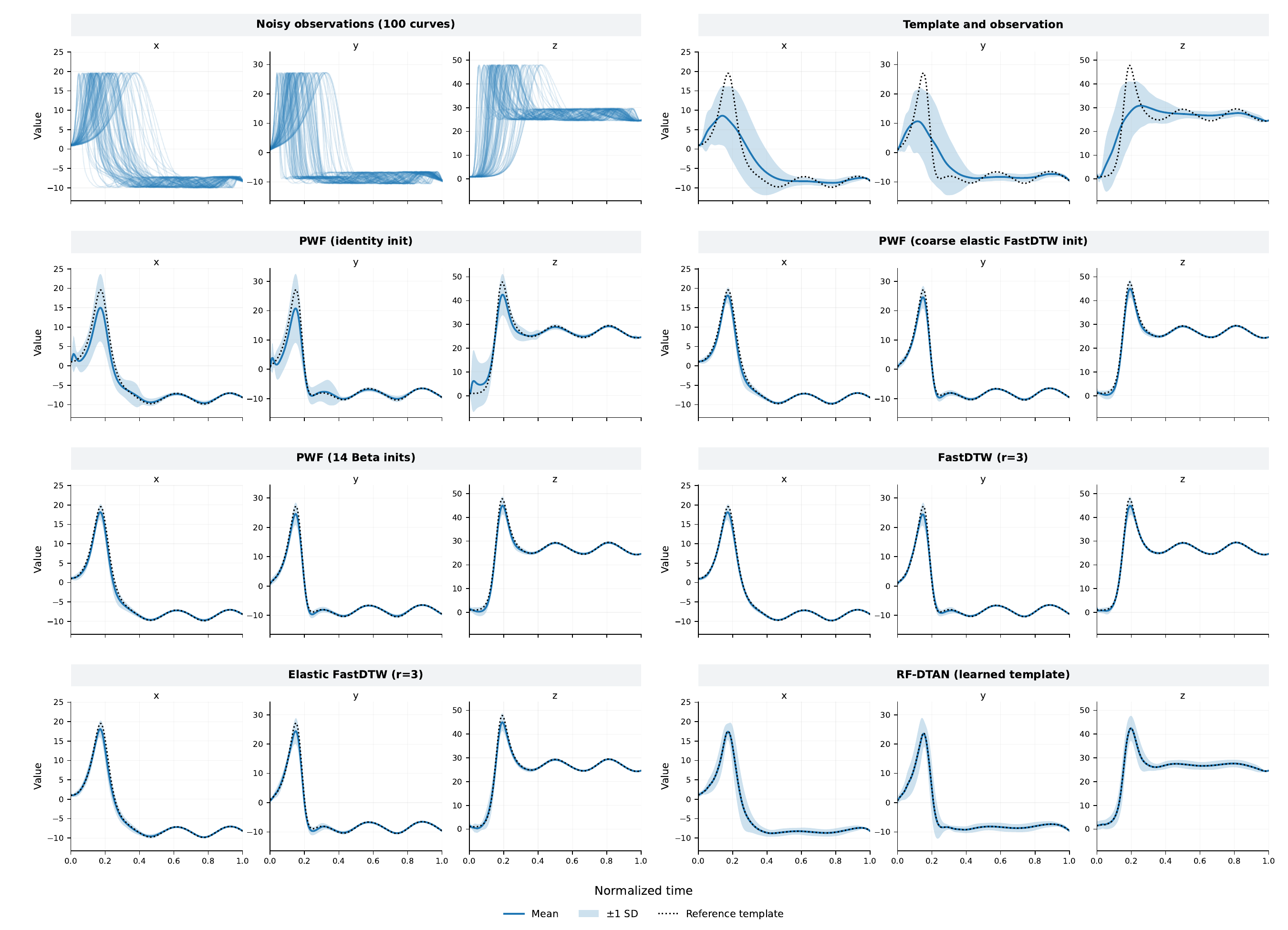}
  \end{subfigure}
\caption{Alignment of 1000 noisy, three-dimensional trajectories simulated from A6.}
\label{fig:A6_lorenz_mean_sd_revised}
\end{figure*}
\cref{fig:A2_peaks_mean_sd_revised}-\cref{fig:A6_lorenz_mean_sd_revised} show the alignment results for scenarios A2-A6. Across all scenarios, PWF recovers the main structure of the template and substantially reduces the phase variability present before alignment. The A6 scenario further demonstrates PWF's ability to align multidimensional functional data. Initialization has a clear effect. With the identity initialization, PWF often converges to suboptimal warps, which attenuates peaks and leaves wider bands, as in A2 and A3. Using 14 Beta initializations consistently gives the best PWF results, closely matching the template and performing comparably to the elastic FastDTW with $r=3$. PWF consistently outperforms  SrvfNet, SrvfRegNet and RF-DTAN: both SrvfNet and RF-DTAN damp the template features, and SrvfRegNet frequently fails to align large parts of the curves. The main limitation of PWF appears in A3, where the template contains sharp, narrow, repeated peaks. Even with multiple initializations, the tallest peaks remain attenuated and the bands stay wide. This suggests that a single smooth global warp has difficulty with many closely spaced local features, unlike elastic FastDTW with $r=3$, which aligns them more accurately. Euclidean  FastDTW produces the tightest bands in every scenario, but as discussed above, this reflects pinching rather then better registration.

\begin{figure*}
  \centering
  \begin{subfigure}{\linewidth}
    \includegraphics[width=\linewidth]{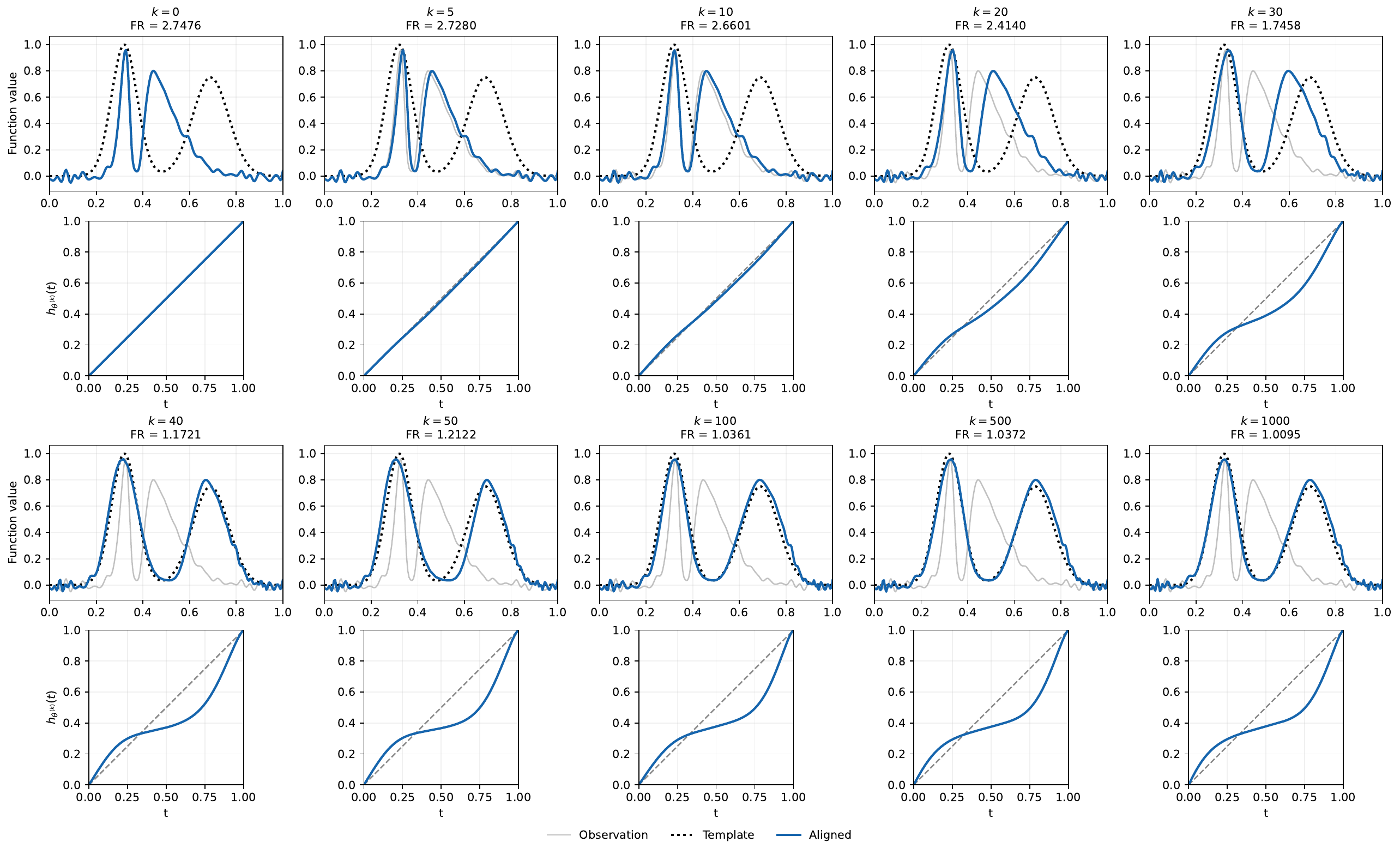}
  \end{subfigure}
\caption{Snapshots of the PWF optimization trajectory for a sample observation from scenario A2, initialized at the identity warp ($\alpha=\beta=1$), at selected iterations \(k\). For each $k$, the upper panel shows the observation (grey), the template (dotted) and the aligned function (blue), and the lower panel shows the current warping function  \(h_{\theta^{(k)}}(t)\), with the identity as a dashed line. FR denotes the Fisher-Rao distance between the template and the aligned function.}
\label{fig:A2_pwf_identity_convergence}
\end{figure*}
\cref{fig:A2_pwf_identity_convergence} shows the optimization trajectory of PWF  initialized at the identify warp $(\alpha=\beta=1)$. The warp changes little during the first 10 iteration, then rapidly develops an S-shape between iterations 10 and 40, moving the second peak toward its target and reducing the Fisher-Rao distance from 2.66 to 1.17. After about 100 iterations, the warp stabilizes and further iterations yield only marginal improvements. \cref{fig:A2_pwf_identity_convergence} illustrates that, for curves with a simple shape, PWF can align the peaks from the identity initialization within a few dozen iterations, without pinching.

\subsection{Details on cluster analysis} \label{cluster_analysis_supplement}

\begin{table}[!ht]
\centering
\caption{Characteristics of the 12 datasets used to evaluate clustering performance. Here, $n$ is the sample size, $T$ is the number of time points per sequence, $p$ is the input dimension, and $K$ is the known number of classes.}
\label{tab:arm-b-datasets}
\resizebox{0.85\linewidth}{!}{%
\begin{tabular}{lcccc}
\toprule
\textbf{Dataset} & $n$ & $T$ & $p$ & $K$ \\
\midrule
\multicolumn{5}{l}{\emph{Scalar functions}} \\
CBF & 300 & 128 & 1 & 3  \\
Crop & 24000 & 46 & 1 & 24  \\
ECG200 & 200 & 96  & 1 & 2  \\
Plane & 210  & 144 & 1  & 7 \\
SwedishLeaf & 1125 & 128 & 1 & 15 \\
Symbols & 1020 & 398 & 1 & 6  \\
Trace & 200 & 275 & 1 & 4  \\
\multicolumn{5}{l}{\emph{Multi-dimensional functions}} \\
BMHAD & 659  & 300 & 129  & 11 \\
ERing & 300 & 65  & 4  & 6 \\
NATOPS & 360 & 51  & 24 & 6 \\
RacketSports & 303 & 30  & 6  & 4 \\
UWaveGestureLibrary & 440 & 315 & 3  & 8 \\
\bottomrule
\end{tabular}%
}
\end{table}

For each dataset, we create a single class-stratified training-validation-test split in a \(70{:}15{:}15\) ratio using random seed 42. All methods use the same saved split. We tune hyperparameters by randomly sampling up to 20 configurations per model from a predefined search space and selecting the configuration with the lowest validation reconstruction loss.

\paragraph{ShapeNet.}
We select hyperparameters by minimizing the Fisher-Rao reconstruction loss. Before training, the functional data are standardized using a single mean and population standard deviation computed from the training set. Each candidate architecture is specified by an ordered list of channel widths, \([p, q_1, q_2, s, d_1, d_2, d_3, d_4, p]\), where \(p\) is the number of input and reconstructed output channels, \(q_1\) is the number of nodes in the weight-orbit layer, \(q_2\) is the width of the subsequent encoder layer, \(s\) is the embedding dimension, and \(d_1, \ldots, d_4\) are the decoder widths. The selected  architectures vary across datasets. Across the search space, \(q_1 \le 32\), \(q_2 \le 32\), and \(s \in \{4, 8, 16\}\). In the decoder, a first layer expands the embedding to width \(d_1 \le 32\), followed by three functional layers with widths \(d_2 \le 64\), \(d_3 \le 32\), and \(d_4 \le 32\).

We also tune the following hyperparameters, with search ranges chosen per dataset: the numbers of basis functions (for the functional weights); the batch size; the network learning rate (approximately \(10^{-3}\) to \(1.5\times10^{-2}\)); the learning rate and number of iterations (25 to 100) for the elastic-registration optimizer; the number of online updates to the initialization-amortizer per batch (15 to 25); the coefficients of the regularization on the functional weights (\(10^{-4}\) to \(10^{-2}\)); and the number of training epochs (50, 100, 150, or 200). In all cases, a step scheduler multiplies the learning rate by 0.75 every 20 epochs.

\paragraph{FAE.}
The functional autoencoder based on \cite{hsieh2021functional} has a nonlinear functional-inner-product encoder and a linear vector-to-function decoder. Functional weights are represented on the observation grid; B-spline or Fourier functions are used for their initialization. The model is trained by minimizing the integrated squared reconstruction loss, and configurations are selected by the same loss on the validation set. The latent dimension is fixed at 20. The remaining hyperparameters are chosen per dataset. The initialization basis (B-spline or Fourier) is chosen to suit the data, with 15, 20, or 25 basis functions. The Adam learning rate for the functional weights is varied over dataset-specific ranges within \(10^{-3}\) to \(1.5\times10^{-2}\), and training runs for 200, 500, or 1,000 epochs.

\paragraph{VANO.}
VANO \cite{seidman2023variational} is trained by minimizing a reconstruction loss plus a Kullback-Leibler divergence term weighted by \(\beta\). The encoder and decoder each have three hidden layers of width 128 with GELU activations, and the variational objective is estimated with 16 Monte Carlo samples. Models are trained for 200 epochs with Adam and an exponentially decaying learning rate. For each dataset, we sample 20 configurations, varying four hyperparameters. The batch size and latent dimension are drawn from dataset-specific sets. The learning rate is sampled log-uniformly from \(3\times10^{-4}\) to \(10^{-2}\), and \(\beta\) is sampled log-uniformly from either \([10^{-6}, 10^{-5}]\) or \([10^{-5}, 10^{-3}]\), depending on the dataset.

\paragraph{FAEclust.}
FAEclust \cite{singh2025NeurIPS} handles phase variation through its own convex-clustering technique. We therefore evaluate this built-in clustering procedure rather than fitting K-means or a GMM to its embeddings. The neighborhood graph used by the convex objective function is built from the training observations only. We set the number of nearest neighbors \(m\) to the larger of two estimates: one based on average distances and one based on graph connectivity. The connectivity-based search is capped at 20 neighbors, but \(m\) can exceed 20 when the average-distance estimate is larger. Graph distances are computed with the elastic-fast distance. To speed up this calculation, curves with more than 100 sampling points are down-sampled to 100 points; the down-sampled curves are used only for graph construction, and the autoencoder is trained on the full curves.

Training consists of reconstruction pretraining followed by joint training with the clustering, encoder, and decoder penalties, weighted by \(\lambda_c\), \(\lambda_e\), and \(\lambda_d\), respectively. Across datasets, the search ranges are 0.05 to 5.0 for \(\lambda_c\), 0.01 to 1.0 for \(\lambda_e\), and 0.01 to 0.1 for \(\lambda_d\). The channel widths, basis type, and number of basis functions are chosen per dataset, and training runs for 100, 150, or 200 epochs. As for the other methods, the configuration is selected by validation reconstruction loss. We also report results under selection by validation ARI, as a label-guided sensitivity analysis.

\cref{tab:arm-b-ami-full,tab:arm-b-ari} report the mean and standard deviation of AMI and ARI, respectively. For completeness, \cref{tab:arm-b-ami-full} also includes the mean AMI values reported in \cref{tab:arm-b-ami}. ShapeNet leads on 8 of the 12 datasets. ShapeNet  achieves perfect clustering on Plane under both clustering methods and strong performance on Symbols, with AMI of 0.930 and ARI of up to 0.927. On BMHAD, ShapeNet (GMM) improves over the strongest baseline from 0.864 to 0.908 in AMI and from 0.741 to 0.839 in ARI. It also achieves the highest scores on NATOPS and RS. ShapeNet falls behind the strongest baseline on Crop, Trace, ERing, and UWGL. The largest gaps occur on ERing and UWGL: VANO (K-means) achieves AMI/ARI of 0.952/0.940 on ERing, compared with ShapeNet’s best scores of 0.635/0.528; FAE (K-means) achieves 0.736/0.656 on UWGL, compared with ShapeNet's 0.308/0.173. FAEcl. performs best on Trace.  ShapeNet's reported standard deviations are generally small, especially on scalar-function datasets, although BMHAD shows greater variability, with an ARI standard deviation of 0.093 for GMM. 
\begin{table*}[!ht]
\centering
\caption{Mean AMI $\pm$ standard deviation over 20 random initializations.}
\label{tab:arm-b-ami-full}
\resizebox{\linewidth}{!}{%
\setlength{\tabcolsep}{5pt}
\begin{tabular}{lccccccc}
\toprule
& \multicolumn{2}{c}{FAE} & \multicolumn{2}{c}{VANO} & FAEclust & \multicolumn{2}{c}{ShapeNet} \\
\cmidrule(lr){2-3} \cmidrule(lr){4-5} \cmidrule(lr){7-8}
\textbf{Data} & K-means & GMM & K-means & GMM &  & K-means & GMM \\
\midrule
\multicolumn{8}{l}{\emph{Scalar functions}} \\
CBF & 0.514$\pm$0.002 & 0.386$\pm$0.000 & 0.380$\pm$0.003 & 0.382$\pm$0.035 & 0.310 & 0.563$\pm$0.000 & \textbf{0.570$\pm$0.000} \\
Crop & 0.533$\pm$0.002 & 0.497$\pm$0.002 & 0.509$\pm$0.004 & 0.564$\pm$0.005 & -- & 0.418$\pm$0.003 & 0.436$\pm$0.006 \\
ECG200 & 0.214$\pm$0.000 & \textbf{0.464$\pm$0.000} & 0.139$\pm$0.030 & 0.253$\pm$0.000 & 0.278 & \textbf{0.464$\pm$0.000} & 0.411$\pm$0.000 \\
Plane & 0.931$\pm$0.000 & 0.931$\pm$0.000 & 0.870$\pm$0.000 & 0.837$\pm$0.048 & 0.931 & \textbf{1.000$\pm$0.000} & \textbf{1.000$\pm$0.000} \\
SL & 0.428$\pm$0.020 & 0.551$\pm$0.018 & 0.444$\pm$0.014 & 0.534$\pm$0.023 & 0.192 & 0.556$\pm$0.010 & \textbf{0.602$\pm$0.012} \\
Symbols & 0.797$\pm$0.002 & 0.806$\pm$0.005 & 0.787$\pm$0.001& 0.575$\pm$0.017 & 0.855 & \textbf{0.930$\pm$0.000} & \textbf{0.930$\pm$0.000} \\
Trace & 0.444$\pm$0.000 & 0.458$\pm$0.000 & 0.486$\pm$0.000 & 0.439$\pm$0.031 & \textbf{0.736} & 0.608$\pm$0.000 & 0.591$\pm$0.000 \\
\addlinespace
\multicolumn{8}{l}{\emph{Multi-dimensional functions}} \\
BMHAD & 0.780$\pm$0.011 & 0.719$\pm$0.036 & 0.864$\pm$0.031 & 0.786$\pm$0.031 & 0.312 & 0.848$\pm$0.013 & \textbf{0.908$\pm$0.039} \\
ERing & 0.596$\pm$0.005 & 0.614$\pm$0.016 & \textbf{0.952$\pm$0.000} & 0.931$\pm$0.012 & 0.647 & 0.596$\pm$0.006 & 0.635$\pm$0.003 \\
NATOPS & 0.483$\pm$0.016 & 0.531$\pm$0.043 & 0.702$\pm$0.000& 0.756$\pm$0.000 & 0.464 & 0.714$\pm$0.000 & \textbf{0.807$\pm$0.046} \\
RS & 0.173$\pm$0.031 & 0.270$\pm$0.082 & 0.215$\pm$0.000 & 0.248$\pm$0.000& 0.260 & 0.246$\pm$0.000 & \textbf{0.374$\pm$0.003} \\
UWGL & \textbf{0.736$\pm$0.014} & 0.457$\pm$0.014 & 0.598$\pm$0.052 & 0.628$\pm$0.014& 0.243 & 0.256$\pm$0.011 & 0.308$\pm$0.038 \\
\bottomrule
\end{tabular}%
}
\end{table*}

\begin{table*}[!ht]
\centering
\caption{Mean ARI $\pm$ standard deviation over 20 random initializations}
\label{tab:arm-b-ari}
\resizebox{\linewidth}{!}{%
\setlength{\tabcolsep}{5pt}
\begin{tabular}{lccccccc}
\toprule
& \multicolumn{2}{c}{FAE} & \multicolumn{2}{c}{VANO} & FAEclust & \multicolumn{2}{c}{ShapeNet} \\
\cmidrule(lr){2-3} \cmidrule(lr){4-5} \cmidrule(lr){7-8}
\textbf{Data} & K-means & GMM & K-means & GMM & & K-means & GMM \\
\midrule
\multicolumn{8}{l}{\emph{Scalar functions}} \\
CBF & 0.471$\pm$0.003 & 0.281$\pm$0.000 & 0.377$\pm$0.004 & 0.344$\pm$0.007 & 0.241 & 0.563$\pm$0.000 & \textbf{0.586$\pm$0.000} \\
Crop & 0.305$\pm$0.005 & 0.269$\pm$0.009 & 0.275$\pm$0.008 & 0.278$\pm$0.010 & -- & 0.190$\pm$0.002 & 0.194$\pm$0.006 \\
ECG200 & 0.319$\pm$0.000 & \textbf{0.427$\pm$0.000} & 0.083$\pm$0.048 & 0.075$\pm$0.000 & 0.374 & \textbf{0.427$\pm$0.000} & 0.340$\pm$0.000 \\
Plane & 0.860$\pm$0.000 & 0.860$\pm$0.000 & 0.794$\pm$0.000 & 0.777$\pm$0.063 & 0.860 & \textbf{1.000$\pm$0.000} & \textbf{1.000$\pm$0.000} \\
SL & 0.229$\pm$0.022 & 0.377$\pm$0.018 & 0.254$\pm$0.014 & 0.346$\pm$0.022 & 0.038 & 0.345$\pm$0.006 & \textbf{0.413$\pm$0.018} \\
Symbols & 0.672$\pm$0.001 & 0.668$\pm$0.008 & 0.666$\pm$0.001 & 0.444$\pm$0.026 & 0.794 & \textbf{0.927$\pm$0.000} & 0.924$\pm$0.000 \\
Trace & 0.304$\pm$0.000 & 0.337$\pm$0.000 & 0.386$\pm$0.000 & 0.347$\pm$0.045 & \textbf{0.671} & 0.499$\pm$0.000 & 0.475$\pm$0.000 \\
\addlinespace
\multicolumn{8}{l}{\emph{Multi-dimensional functions}} \\
BMHAD & 0.517$\pm$0.010 & 0.520$\pm$0.066 & 0.741$\pm$0.055 & 0.643$\pm$0.025 & 0.123 & 0.723$\pm$0.046 & \textbf{0.839$\pm$0.093} \\
ERing & 0.501$\pm$0.010& 0.500$\pm$0.012 & \textbf{0.940$\pm$0.000} & 0.900$\pm$0.023 & 0.444 & 0.495$\pm$0.010 & 0.528$\pm$0.004 \\
NATOPS & 0.293$\pm$0.021 & 0.374$\pm$0.048 & 0.624$\pm$0.000 & 0.673$\pm$0.000 & 0.269 & 0.554$\pm$0.000 & \textbf{0.713$\pm$0.000} \\
RS & 0.143$\pm$0.031 & 0.209$\pm$0.064 & 0.094$\pm$0.000 & 0.124$\pm$0.000 & 0.188 & 0.162$\pm$0.000 & \textbf{0.283$\pm$0.000} \\
UWGL & \textbf{0.656$\pm$0.015} & 0.332$\pm$0.014 & 0.498$\pm$0.089 & 0.549$\pm$0.052 & 0.079 & 0.116$\pm$0.006 & 0.173$\pm$0.032 \\
\bottomrule
\end{tabular}%
}
\end{table*}

\subsection{Details on classification analysis} \label{classification_analysis_supplement}
All four classifiers are trained with the Adam optimizer to minimize multiclass cross-entropy, augmented with the method-specific penalties described below. For each dataset, every method is tuned over 10 hyperparameter configurations using the same training, validation, and test sets. No class weighting or resampling is applied. FMLP \cite{rossi2005functional}, AdaFNN \cite{yao2021deep}, and FNN \cite{pmlr-v202-heinrichs23a} are trained for 200 epochs, whereas the number of epochs for ShapeNet-C is selected from \(\{50, 100, 200\}\), following the corresponding search space in the clustering analysis. Crop is handled differently because of its large sample size: all methods are trained for 100 epochs, with the batch size selected from \(\{1024, 1536\}\). For each configuration, we retain the checkpoint with the lowest validation cross-entropy across epochs. We then select the configuration whose retained checkpoint has the lowest validation cross-entropy, breaking ties first by higher validation accuracy, then by fewer trainable parameters, and finally by a fixed deterministic ordering of configurations. The selected checkpoint is evaluated once on the held-out test set.

\paragraph{ShapeNet-C.}
ShapeNet-C is trained to minimize multiclass cross-entropy plus a roughness penalty on its functional weights and biases. Its hyperparameter search space extends that of ShapeNet with larger encoders: up to 32 shape templates in the weight-orbit layer, followed by MLP layers with widths of up to 64 and 128. We sample 20 configurations for each dataset. For multi-dimensional datasets, the search also includes the warp-sharing strategy, choosing between a single warp shared across all channels and a separate warp for each channel.

\paragraph{FMLP.}
The functional multilayer perceptron (FMLP) \cite{rossi2005functional} represents each first-layer functional weight as a linear combination of fixed basis functions. Inner products between these weights and the input curves, computed by quadrature, yield scalar features that pass through a hyperbolic tangent activation and then a classifier with three hidden layers of 128 units each. The basis family is fixed for each dataset to either B-splines or a Fourier basis. We select the number of basis functions from \(\{8, 16, 32\}\), the number of functional units from \(\{16, 32, 64\}\), the dropout rate from \(\{0, 0.1, 0.25\}\), and the weight decay from \(\{0, 10^{-4}\}\), and we sample the learning rate log-uniformly from \([5\times10^{-4}, 2\times10^{-2}]\). For datasets other than Crop, the batch size is selected from \(\{32, 64\}\), except for CBF and ECG200, where it is fixed at 64.

\paragraph{AdaFNN.}
AdaFNN \cite{yao2021deep} learns its basis functions from the data, representing each one with a small neural network. The input curves are projected onto these learned basis functions, and the resulting scores serve as features for the downstream network.  Because the official implementation is designed to regress a scalar response on a single functional covariate, we adapt it to multichannel classification in three ways: the learned basis functions are shared across channels, the scores for all channel-basis pairs are concatenated, and the scalar regression output is replaced by a multiclass classification head. The adaptive layer learns four basis functions, and the resulting scores are passed to an MLP classifier with three hidden layers of width 128. The cross-entropy loss is augmented with an L1 sparsity penalty and a pairwise orthogonality penalty on the basis functions. Nine configurations cover all combinations of orthogonality coefficients \(\{0, 0.5, 1\}\) and sparsity coefficients \(\{0, 1, 2\}\), using basis networks with three hidden layers of width 64. The tenth configuration uses the larger basis network of \cite{yao2021deep}, with four hidden layers of width 256, and sets both penalties to zero. The dropout rate is fixed at \(0.1\), the weight decay at zero, and the learning rate at \(3\times10^{-4}\).

\paragraph{FNN.}
The functional neural network (FNN) \cite{pmlr-v202-heinrichs23a} consists of functional convolution and functional dense layers. Input curves are first smoothed with FNN's local-linear smoother and normalized over the functional domain. They then pass through one or two functional convolution layers with ELU activations, followed by a pooled functional dense layer that produces the class scores. Each convolutional kernel is represented by five Legendre basis functions, and the final dense layer by five Fourier basis functions. The convolutional channel widths are selected from \(\{[40, 20], [5, 10], [3, 12], [20]\}\), and the learning rate is sampled log-uniformly from \([5\times10^{-4}, 2\times10^{-2}]\). Dropout and weight decay are fixed at zero. Functional kernels are evaluated at 25 points on all datasets except Crop, which has only 46 time points; there, kernels are evaluated at 15 points so that every candidate two-layer architecture remains valid.

\paragraph{Significance tests.}
% \begin{table}[!ht]
% \centering
% \caption{Mean paired differences and two-sided Nadeau-Bengio corrected resampled \(t\)-test \(p\)-values for clustering and classification.}
% \label{tab:significance_test}
% \setlength{\tabcolsep}{4pt}
% \begin{tabular}{lrcrc}
% \toprule
% & \multicolumn{2}{c}{Clustering} & \multicolumn{2}{c}{Classification} \\
% \cmidrule(lr){2-3} \cmidrule(lr){4-5}
% \textbf{Data} & \textbf{mean $\Delta$} & \textbf{$t$-test $p$} & \textbf{mean $\Delta$} & \textbf{$t$-test $p$} \\
% \midrule
% \multicolumn{5}{l}{\emph{Scalar functions}} \\
% CBF & -0.183 & 0.152 & -0.009 & 0.076 \\
% Crop & -- & -- & -0.083 & 0.000 \\
% ECG200 & -0.072& 0.535& -0.010& 0.836 \\
% Plane & 0.006 & 0.841 & -0.025 & 0.272 \\
% SL & 0.037 & 0.193 & -0.062 & 0.002 \\
% Symbols & 0.135 & 0.003 & -0.017 & 0.032 \\
% Trace & -0.060 & 0.392 & -0.180 & 0.001 \\

% \addlinespace
% \multicolumn{5}{l}{\emph{Multidimensional functions}} \\
% BMHAD & 0.088 & 0.071 & 0.047 & 0.149 \\
% ERing& -0.254 & 0.000 & -0.039 & 0.276 \\
% NATOPS & 0.143 & 0.005 & -0.011 & 0.731 \\
% RS & 0.036 & 0.502 & -0.078 & 0.127 \\
% UWGL & -0.303 & 0.000 & -0.114 & 0.000 \\
% \bottomrule
% \end{tabular}%
% \end{table}

\begin{table*}[!ht]
\centering
\caption{Mean paired differences over 20 splits (clustering: test AMI; classification: balanced accuracy) with two-sided Nadeau-Bengio corrected resampled $t$-test $p$-values (unadjusted). ShapeNet vs baseline: ShapeNet minus the strongest competitor. $\Delta_A^C$, $\Delta_A^S$: weight-orbit alignment enabled minus disabled. $\Delta_{NA}^C$: full ShapeNet minus a variant without alignment and with $L^2$ instead of Fisher-Rao reconstruction loss. Positive values favour ShapeNet or alignment; -- : not evaluated. }
\label{tab:significance_test}
% \scriptsize
\setlength{\tabcolsep}{4pt}
\begin{tabular}{lcccc|cccccc}
\toprule
& \multicolumn{2}{c}{Clustering}
& \multicolumn{2}{c}{Classification}
& \multicolumn{4}{c}{Clustering}
& \multicolumn{2}{c}{Classification} \\
\cmidrule(lr){2-3}\cmidrule(lr){4-5}
\cmidrule(lr){6-9}\cmidrule(lr){10-11}
& \multicolumn{2}{c}{ShapeNet vs baseline}
& \multicolumn{2}{c}{ShapeNet-C vs baseline}
& \multicolumn{2}{c}{Weight-orbit alignment}
& \multicolumn{2}{c}{Full vs no alignment}
& \multicolumn{2}{c}{Weight-orbit alignment} \\
\cmidrule(lr){2-3}\cmidrule(lr){4-5}
\cmidrule(lr){6-7}\cmidrule(lr){8-9}
\cmidrule(lr){10-11}
\textbf{Data}
& $\boldsymbol{\Delta}$ & \textbf{$p$}
& $\boldsymbol{\Delta}$ & \textbf{$p$}
& $\boldsymbol{\Delta}_{A}^{C}$ & \textbf{$p$}
& $\boldsymbol{\Delta}_{NA}^{C}$ & \textbf{$p$}
& $\boldsymbol{\Delta}_{A}^{S}$ & \textbf{$p$} \\
\midrule
\multicolumn{11}{l}{\emph{Scalar functions}} \\
CBF
& -0.183 & 0.152
& -0.009 & 0.076
& -0.043 & 0.786 & -0.093 & 0.471
& -0.006 & 0.230 \\

Crop
& -- & --
& -0.083 & $<$0.001
& -- & -- & -- & --
& -0.057 & $<$0.001 \\

ECG200
& -0.072 & 0.535
& -0.010 & 0.836
& -0.015 & 0.877 & 0.038 & 0.745
& 0.001 & 0.976 \\

Plane
& 0.006 & 0.841
& -0.025 & 0.272
& -0.004 & 0.933 & 0.069 & 0.030
& -0.023 & 0.256 \\

SL
& 0.037 & 0.193
& -0.062 & 0.002
& -0.027 & 0.420 & 0.078 & 0.017
& -0.017 & 0.302 \\

Symbols
& 0.135 & 0.003
& -0.017 & 0.032
& 0.049 & 0.038 & 0.109 & $<$0.001
& -0.005 & 0.641 \\

Trace
& -0.060 & 0.392
& -0.180 & 0.001
& 0.026 & 0.629 & 0.049 & 0.308
& 0.299 & $<$0.001 \\

\addlinespace
\multicolumn{11}{l}{\emph{Multidimensional functions}} \\
BMHAD
& 0.088 & 0.071
& 0.047 & 0.149
& 0.106 & 0.021 & 0.194 & 0.003
& 0.063 & $<$0.001 \\

ERing
& -0.254 & $<$0.001
& -0.039 & 0.276
& -0.061 & 0.326 & -0.116 & 0.129
& -0.024 & 0.538 \\

NATOPS
& 0.143 & 0.005
& -0.011 & 0.731
& 0.152 & 0.023 & 0.320 & $<$0.001
& 0.194 & $<$0.001 \\

RS
& 0.036 & 0.502
& -0.078 & 0.127
& 0.122 & 0.082 & 0.054 & 0.316
& 0.043 & 0.407 \\

UWGL
& -0.303 & $<$0.001
& -0.114 & $<$0.001
& 0.028 & 0.477 & -0.054 & 0.158
& 0.039 & 0.226 \\
\bottomrule
\end{tabular}
\end{table*}

To assess whether the observed performance differences were stable and statistically significant, we repeated the evaluation on 20 matched stratified train-validation-test splits for each dataset, using the model configurations selected in the original experiments. ShapeNet and its strongest competitors were trained and evaluated on the same split in every repetition. First four columns of \cref{tab:significance_test} report the mean paired difference in test performance and the corresponding Nadeau-Bengio corrected resampled \(t\)-test \(p\)-value for both clustering and classification. Positive differences favour ShapeNet. This correction accounts for dependence among overlapping splits, while Holm’s procedure is subsequently used to assess study-wide evidence across datasets.

The repeated-split analysis provides the strongest support for ShapeNet in unsupervised representation learning. ShapeNet achieved a higher mean test AMI than the strongest competitor on 6 of the 11 clustering comparisons, with significant improvements of \(0.135\) on Symbols and \(0.143\) on NATOPS. Both advantages remained significant after Holm correction across datasets, with adjusted \(p\)-values of \(0.0234\) and \(0.0377\), respectively. ShapeNet also improved mean AMI by \(0.088\) on BMHAD, although this result was inconclusive. These findings demonstrate that ShapeNet can achieve statistically significant improvements over the strongest competing methods on selected datasets, particularly when learning unsupervised representations.

In classification, ShapeNet-C achieved a higher mean balanced accuracy than the strongest baseline only on BMHAD, but this advantage was not statistically significant. In contrast, the competitor advantages on Crop, SwedishLeaf, Trace, and UWaveGestureLibrary remained significant after Holm correction. The difference on Symbols did not remain significant after this adjustment. The classification results suggest that explicit phase invariance offers a less consistent advantage in supervised settings, potentially because discriminative training can already learn representations that tolerate some temporal variability. Note that these analyses assess the stability of fixed, previously selected configurations rather than repeating the complete hyperparameter-selection procedure within every resplit.

\paragraph{Alignment ablation.} ShapeNet aligns functions in two places: the weight-orbit layer aligns each input to the learned shape template before taking inner products, and the Fisher-Rao reconstruction loss aligns reconstructions to their targets. We assess these with two ablations. The first disables weight-orbit alignment but keeps the Fisher-Rao loss, isolating encoder alignment ($\Delta_A^C$). The second also replaces the
Fisher-Rao loss with the $L^2$ loss, removing both components ($\Delta_{NA}^C$).

Weight-orbit alignment significantly improved clustering AMI on Symbols (\(\Delta_A^C=0.049\), \(p=0.038\)), BMHAD (\(\Delta_A^C=0.106\), \(p=0.021\)) and NATOPS (\(\Delta_A^C=0.152\), \(p=0.023\)) in the corrected resampled tests (\cref{tab:significance_test}). Relative to the joint ablation, the full ShapeNet achieved significant improvements on Plane, SwedishLeaf, Symbols, BMHAD, and NATOPS. These results provide evidence for the contribution of the combined phase-invariant formulation, although the joint comparison does not isolate the individual effects of encoder alignment and reconstruction loss. In classification, weight-orbit alignment significantly improved balanced accuracy on Trace (\(\Delta_A^S=0.299\)), BMHAD (\(0.063\)), and NATOPS (\(0.194\)), while significantly reducing accuracy on Crop (\(-0.057\)). Overall, these results demonstrate that alignment can provide substantial downstream improvements, particularly on selected multidimensional datasets, although its benefits remain dataset dependent.

\subsection{Analysis of neuroimaging data}\label{Analysis of neuroimaging data_supp}

BCI Competition IV Dataset 2a contains four-class motor-imagery EEG from nine subjects performing left-hand, right-hand, both-feet, and tongue imagery. Each subject completed two sessions (T and E) on different days, with 288 trials per session. EEG was recorded from 22 channels at 250\,Hz. We extracted the 2--6\,s interval after trial onset, yielding inputs of $22\times1000$ samples. To assess within-subject cross-session generalization, models were trained and evaluated separately for each subject. Session T was stratified into 75\% training and 25\% validation data, and the complete session E was reserved for testing. Validation loss determined configuration and checkpoint selection. Each reported model was evaluated using 10 matched random seeds per subject.

\begin{figure*}[!ht]
\centering
\includegraphics[width=0.85\linewidth]{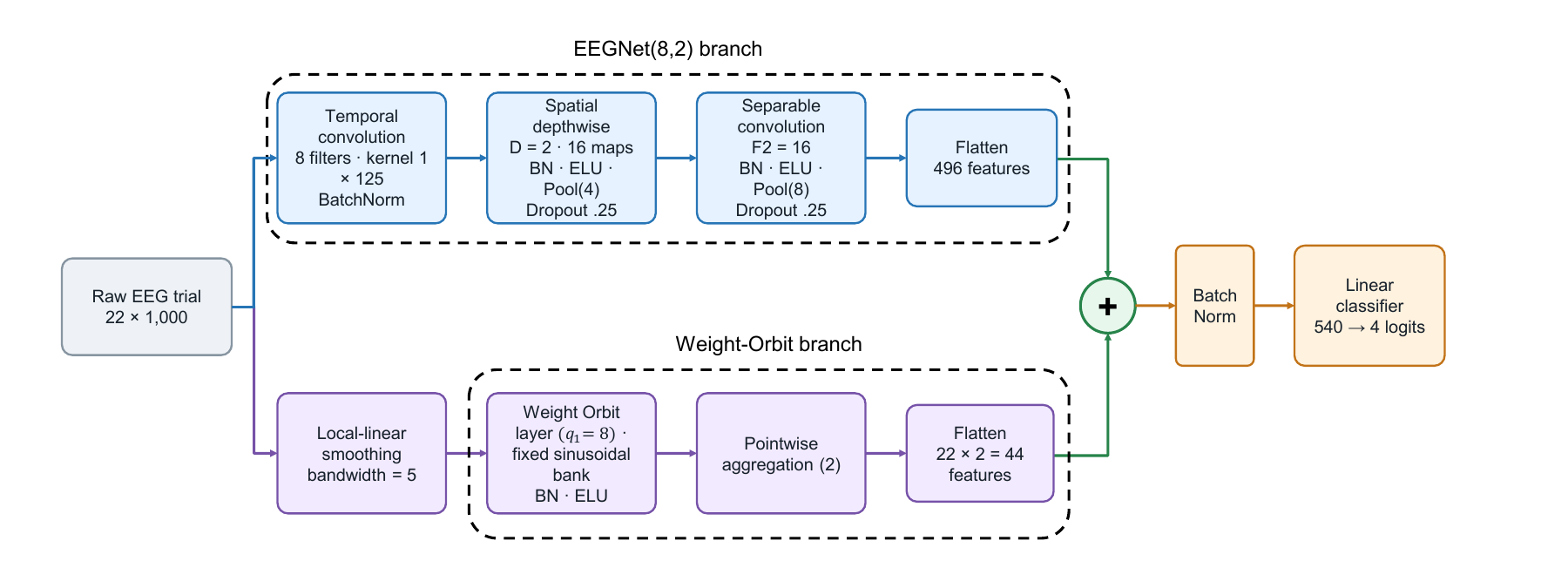}
\caption{ShapeNet-E architecture. EEGNet and weight-orbit features are concatenated, batch-normalized and passed to a linear classifier.}
    \label{fig:shapenet_e}
\end{figure*}

\paragraph{ShapeNet-E architecture.}
ShapeNet-E (\cref{fig:shapenet_e}) combines an EEGNet$(8,2)$ branch with a weight-orbit branch comprising $q_1=8$ nodes in the weight-orbit layer, where each node represents a $d$-dimensional vector ($d$ is the number of electrodes). While EEGNet learns frequency-specific spatial projections across electrodes, the weight-orbit branch is designed to preserve electrode-specific morphological information over the entire recording. To retain electrode-wise shape information, we aggregate the node information separately for each electrode-dimension. Specifically, for each electrode $e$, the aggregated features are given by
\begin{equation*}
    x_{k,e}^{(2)} = \sum_{q=1}^{q_1} a_{k,q} x_{q,e}^{(1)}, \qquad k=1,2,
\end{equation*}
where $a_{k,q}$ are learnable coefficients shared across all electrodes. This produces $2D$ ShapeNet features without mixing information across electrodes, allowing the morphological features extracted from each electrode to contribute directly to the class logits. The ShapeNet features are then flattened and concatenated with the flattened EEGNet features. The resulting joint representation is batch-normalized before being passed to a linear classifier.

\paragraph{Training and hyperparameter selection.}
All models were trained using Adam. EEGNet$(8,2)$ used $F_1=8$, $D=2$, $F_2=16$, a 125-sample temporal kernel, dropout 0.25, batch size 64, and a learning rate $10^{-3}$ for up to 200 epochs with early-stopping patience of 50 epochs. FNN used two functional layers (40 and 20 units), local-linear smoothing (bandwidth 5, derivative bandwidth 10), five Legendre basis functions per functional convolution and a five-function Fourier dense representation. It was  trained for 2,250 sampled mini-batches of size 32 with a  learning rate of $10^{-3}$.  ShapeNet-E used shared warping across electrodes and was trained for 150 epochs with batch size 32 and an initial learning rate of $10^{-2}$, decayed by a factor of 0.9 every 10 epochs. Two fixed sinusoidal frequency banks were considered: $\{1,3,5,7,9,11,13\}\,\mathrm{Hz}$ and $\{4,6,8,10,12,14,16\}\,\mathrm{Hz}$. For each subject, the configuration minimizing session-T validation cross-entropy was retained for evaluation across 10 random seeds. Session E was reserved exclusively for testing.

% \begin{table}[!ht]
% \centering
% \caption{Subject-level balanced accuracy under within-subject cross-session and cross-subject evaluation. Mean and standard deviation are computed across the nine subjects.}
% \label{tab:eeg_results}
% \begin{tabular}{ccccc}
% \toprule
% & \multicolumn{2}{c}{Cross-session} & \multicolumn{2}{c}{Cross-subject} \\
% \cmidrule(lr){2-3}\cmidrule(lr){4-5}
% Sub. & ShapeNet-E & EEGNet & ShapeNet-E & EEGNet\\
% \midrule
% A01 & 0.790 & 0.719 & 0.606 & {0.632} \\
% A02 & {0.547} & 0.525 & 0.458 & {0.514} \\
% A03 & {0.897} & 0.810 & {0.793} & 0.710 \\
% A04 & {0.637} & 0.522 & 0.422 & {0.446} \\
% A05 & 0.651 & {0.667} & 0.405 & 0.411 \\
% A06 & {0.578} & 0.545 & 0.438 & {0.505} \\
% A07 & {0.723} & 0.695 & 0.670 & {0.691} \\
% A08 & {0.772} & 0.735 & 0.578 & {0.667} \\
% A09 & {0.801} & 0.723 & 0.559 & {0.606} \\
% \midrule
% mean & {0.711} & 0.660 & 0.548 & {0.576}\\
% sd & {0.115} & 0.105 & 0.131 & {0.110}\\
% \bottomrule
% \end{tabular}
% \end{table}

\begin{table}[!ht]
\centering
\caption{Subject-level balanced accuracy. Mean and standard deviation are computed across the nine subjects.}
\label{tab:eeg_results}
\begin{tabular}{cccc}
\toprule
Sub. & ShapeNet-E & EEGNet(8,2) & FNN \\
\midrule
A01 & 0.806 & 0.719 & 0.544\\
A02 & {0.557} & 0.525 & 0.363 \\
A03 & {0.904} & 0.810 & 0.537\\
A04 & {0.626} & 0.522 & 0.336\\
A05 & 0.678 & {0.667} & 0.537\\
A06 & {0.570} & 0.545 & 0.383 \\
A07 & {0.754} & 0.695 & 0.552 \\
A08 & {0.763} & 0.735 & 0.506\\
A09 & {0.799} & 0.723 & 0.514\\
\midrule
mean & {0.717} & 0.660 & 0.475\\
sd & {0.117} & 0.105 & 0.088\\
\bottomrule
\end{tabular}
\end{table}
\Cref{tab:eeg_results} reports subject-level balanced accuracy  under the within-subject cross-session evaluation protocol. ShapeNet-E achieved the highest mean balanced accuracy ($0.717$), outperforming EEGNet and FNN by $0.057$ and $0.242$, respectively. Notably, ShapeNet-E consistently outperformed both baselines across all nine subjects, with improvements over EEGNet ranging from $0.011$ to $0.104$. Substantial improvements were observed for subjects with both relatively low (e.g., A04) and high (e.g., A03) baseline accuracies, indicating that the benefits are not restricted to subjects for whom EEGNet performs poorly. These results highlight the potential of incorporating morphological features into conventional EEG architectures for cross-session generalization.

To separate the contribution of the weight-orbit branch from temporal alignment and model capacity, we evaluated ShapeNet-E with alignment disabled and used a wider EEGNet$(16,2)$ with \(F_2=32\) as a capacity control. ShapeNet-E contains 6,612 trainable parameters: 1,944 in the EEGNet feature extractor, 2,504 in the weight-orbit branch, and 2,164 in the fusion classifier.  EEGNet$(16,2)$ contains 8,372 parameters and was trained and evaluated using the same procedure as EEGNet$(8,2)$. 

\begin{table}[t]
\centering
\caption{Cross-session ablations. Balanced accuracy is reported as mean $\pm$ standard deviation across nine subject-level means.}
\label{tab:eeg_ablation}
\setlength{\tabcolsep}{2pt}
\begin{tabular}{lcc}
\toprule
Model & Parameters & Balanced accuracy \\
\midrule
EEGNet$(8,2)$ & 3,932 & $0.660 \pm 0.105$ \\
EEGNet$(16,2)$ & 8,372 & $0.679 \pm 0.113$ \\
ShapeNet-E, align. disabled & 6,612 & $0.681 \pm 0.123$ \\
ShapeNet-E & 6,612 & $0.717 \pm 0.117$ \\
\bottomrule
\end{tabular}
\end{table}

Table \ref{tab:eeg_ablation} shows that increasing EEGNet width improved balanced accuracy from $0.660$ to $0.679$. Nevertheless, ShapeNet-E achieved $0.717$ with fewer trainable parameters than EEGNet$(16,2)$, outperforming it by 0.039 on average across the nine subjects (paired 95\% CI $[0.020,\,0.058]$; paired $t$-test, $p=0.002$; Wilcoxon test, $p=0.008$).  Disabling temporal alignment while retaining the same architecture and parameter count reduced mean balanced accuracy by $0.037$ (paired 95\% CI $[0.026,\,0.047]$). The reduction was significant under both the paired $t$-test ($p < 0.001$) and the Wilcoxon signed-rank test ($p=0.004$). Under this matched cross-session comparison, the results provide evidence that temporal alignment contributes to the improved performance of ShapeNet-E.

% To separate the contribution of the weight-orbit branch, we compared ShapeNet-E with EEGNet$(8,2)$ and EEGNet$(16,2)$ with \(F_2=32\) as a capacity control. ShapeNet-E contains 6,836 trainable parameters: 1,944 in the EEGNet feature extractor, 2,840 in the weight-orbit branch, and 2,052 in the fusion classifier.  EEGNet$(16,2)$ contains 8,372 parameters and was trained and evaluated using the same procedure as EEGNet$(8,2)$.

% \begin{table}[t]
% \centering
% \caption{Within-subject cross-session balanced accuracy (mean $\pm$ sd over nine subjects). EEGNet(16,\,2) is a capacity control. }
% \label{tab:eeg_ablation}
% \setlength{\tabcolsep}{2pt}
% \begin{tabular}{lcc}
% \toprule
% Model & Parameters & Balanced accuracy \\
% \midrule
% EEGNet$(8,2)$ & 3,932 & $0.660 \pm 0.105$ \\
% EEGNet$(16,2)$ & 8,372 & $0.679 \pm 0.113$ \\
% ShapeNet-E& 6,836 & $0.711 \pm 0.115$ \\
% \bottomrule
% \end{tabular}
% \end{table}

\end{document}